\documentclass[a4paper,fleqn]{cas-sc}

\usepackage[authoryear,longnamesfirst]{natbib}
\usepackage{multirow}
\usepackage{amsmath}
\usepackage{amsfonts}
\usepackage{subfigure}
\usepackage{lineno}

\def\tsc#1{\csdef{#1}{\textsc{\lowercase{#1}}\xspace}}
\tsc{WGM}
\tsc{QE}

\begin{document}
\let\WriteBookmarks\relax
\def\floatpagepagefraction{1}
\def\textpagefraction{.001}

\shorttitle{GOMS data for NTFE}    

\shortauthors{Z.H, Z.Z, M.M, W.M}  

\title [mode = title]{Network-Wide Traffic Flow Estimation Across Multiple Cities with Global Open Multi-Source Data: A Large-Scale Case Study in Europe and North America}  


%

\affiliation[PolyUCEE]{organization={Department of Civil and Environmental Engineering, The Hong Kong Polytechnic University},
            addressline={Kowloon}, 
            city={Hong Kong SAR},
            country={China}
            }

\affiliation[NYUADDE]{organization={Division of Engineering, New York University Abu Dhabi, Abu Dhabi, United Arab Emirates},
            city={Abu Dhabi},
            country={United Arab Emirates}
            }            
\author[1]{Zijian Hu}[]
\ead{zijian.hu@connect.polyu.hk}

\author[1]{Zhenjie Zheng}[]
\ead{zzj17.zheng@polyu.edu.hk}

\author[2]{Monica Menendez}[]
\ead{monica.menendez@nyu.edu}

\author[1]{Wei Ma}[]
\ead{wei.w.ma@polyu.edu.hk}


\begin{abstract}
Network-wide traffic flow, which captures dynamic traffic volume on each link of a general network, is fundamental to smart mobility applications.
However, the observed traffic flow from sensors is usually limited across the entire network due to the associated high installation and maintenance costs. To address this issue, existing research uses various supplementary data sources to compensate for insufficient sensor coverage and estimate the unobserved traffic flow. Although these studies have shown promising results, the inconsistent availability and quality of supplementary data across cities make their methods typically face a trade-off challenge between accuracy and generality. In this research, we first time advocate using the Global Open Multi-Source (GOMS) data within an advanced deep learning framework to break the trade-off. The GOMS data primarily encompass geographical and demographic information, including road topology, building footprints, and population density, which can be consistently collected across cities. More importantly, these GOMS data are either causes or consequences of transportation activities, thereby creating opportunities for accurate network-wide flow estimation. Furthermore, we use map images to represent GOMS data, instead of traditional tabular formats, to capture richer and more comprehensive geographical and demographic information. To address multi-source data fusion, we develop an attention-based graph neural network that effectively extracts and synthesizes information from GOMS maps while simultaneously capturing spatiotemporal traffic dynamics from observed traffic data. A large-scale case study across 15 cities in Europe and North America was conducted. The results demonstrate stable and satisfactory estimation accuracy across these cities, which suggests that the trade-off challenge can be successfully addressed using our approach.
\end{abstract}


\begin{highlights}
    \item We advocate the use of Global Open Multi-Source (GOMS) data to estimate network-wide traffic flow across cities.
    
    \item Map images, rather than tabular data, are used to characterize GOMS data more comprehensively.
    
    \item Advanced attention-based graph neural network modules are developed to enable multi-source data fusion.
    
    \item A large-scale case study across 15 cities in Europe and North America is conducted.

\end{highlights}

\maketitle

\section{Introduction}\label{sec:intro}
Network-wide traffic flow, which represents the dynamic traffic volumes on each link of a road network, is fundamental to smart mobility applications, such as traffic signal control \citep{signal_control_flow}, travel time estimation \citep{travel_time_flow} and transportation planning \citep{pelletier2011smart}. In many cities, traffic flow is currently collected using stationary sensors, including loop detectors \citep{detector_flow} and surveillance cameras \citep{camera_flow}. However, the installed stationary sensors are usually insufficient to cover the entire network due to the associated high installation and maintenance costs \citep{cellphone_flow, google_flow}. Consequently, directly acquiring network-wide flow from sensors remains challenging in cities. In view of this, it is of great importance to develop network-wide traffic flow estimation (NTFE) methods with a limited number of sensors.

Existing research on NTFE methods tends to utilize supplementary data sources to make up for the insufficient sensor coverage by exploiting the latent correlation between supplementary data and traffic flow. However, to the best of our knowledge, these NTFE methods typically face the challenge of a trade-off between accuracy and generality. Specifically, the inclusion of more supplementary data affects: i) accuracy: higher accuracy in the NTFE method could be achieved; ii) generality: fewer cities can utilize the NTFE method due to the lack of the required data. Conversely, seeking to enhance the generality also compromises its accuracy. As a result, without addressing the trade-off challenge, public agencies in each city are compelled to devise their own NTFE methods based on localized experiences and expertise. This is inefficient, laborious, and inherently unreliable, resulting in inconsistent and potentially flawed outcomes across cities. More critically, traffic data within each city can only be used for that specific city, and this hinders effective usage of the data. Overall, how to develop NTFE methods with both satisfactory accuracy and generality for multiple cities has not been well explored in the literature.

We broadly categorize the existing studies on NTFE methods into two streams from a data perspective. The first stream mainly relies on Floating Car Data (FCD) and observed data from stationary sensors that are generally available across cities. While NTFE can be successfully achieved by leveraging the relationship between FCD and traffic flow, the main challenge in this stream of research stems from the restricted information provided by FCD alone. Some studies \citep{est_MFD_1,est_MFD_2,est_MFD_3} aim to calibrate the Macroscopic Fundamental Diagram (MFD) using FCD and observed data from sensors, which establishes a fundamental relationship between space-mean flow, density, and network speed for homogeneous regions \citep{MFD}. Subsequently, unobserved traffic flow can be successfully estimated based on the MFD, as long as the unobserved road segments belong to previously calibrated homogeneous regions. However, the MFD-based methods struggle to capture the temporal dependencies inherent in traffic flow dynamics, which limits their accuracy. The challenge of determining homogeneous regions with low-penetration FCD further complicates their applications in the real world. To address these limitations, researchers have increasingly explored advanced machine learning methods, such as Graph Neural Network (GNN), Recurrent Neural Network (RNN), and other deep learning methods \citep{tgc_rnn_flow,sgmc_flow,GNN_new_flow}, to estimate the network-wide flow by exploiting the spatial-temporal relationship between link-level traffic flow and speed. Although the spatial-temporal relationship can be well-encoded in the neural networks, they may still demonstrate unsatisfactory accuracy, particularly for roads not included in the training set. Essentially, the limited information in the FCD is insufficient to capture traffic flow dynamics on unobserved roads and leads to accuracy issues.

The second stream tends to utilize a wide range of supplementary data tailored to a specific city to achieve the NTFE, which may achieve satisfactory accuracy but suffers from generality issues. Early studies in this stream often employ the Origin-Destination (OD) data, land use, and household surveys in each city for NTFE using the traditional four-step models \citep{daganzo1997fundamentals, lam2001activity, vuchic2007urban}. However, considerable effort and cost are required to collect the data. Moreover, it is not trivial to implement the four-step model even if the data is ready. Recently, there has been an emerging trend towards utilizing other data sources, such as cellphone counts \citep{cellular_flow,cellphone_flow}, License Plate Recognition (LPR) data \citep{zhan2020link, gps_lpr_flow_1} and surveillance cameras \citep{camera_flow}, to improve the NTFE accuracy by using FCD alone. The rapid development of data science and deep learning has further expanded the analytical tools, enabling the incorporation of sophisticated inputs such as Point of Interest (POI) data, network topology, and granular road features for NTFE. However, the inconsistent availability of these data sources prevents their methods from being applied across cities.

To break the aforementioned trade-off between accuracy and generality, we advocate the use of Global Open Multi-Source (GOMS) data for NTFE. The GOMS data mainly refers to the globally and publicly available geographical and demographical information, including elements such as road topology, building footprints, and population density \citep{maplur} that contribute to NTFE. Importantly, these GOMS data can significantly enhance the NTFE accuracy, as the network-wide traffic flow data are either the cause or the consequence of the urban activities recorded in the GOMS data. Preliminary research utilizing GOMS data has also shown promising results in estimating land-use patterns \citep{maplur} and air pollution \citep{ozone_lur}. In the context of NTFE, the employed GOMS data should meet two requirements: 1) The GOMS data should be widely available in multiple cities and therefore can address the generality issue; 2) The GOMS data should contain sufficient information to contribute to a more accurate NTFE. Moreover, it is foreseeable that massive GOMS data will emerge with the advent of smart sensor technologies to further enhance the estimation accuracy. 

Compared to traditional tabular GOMS data, GOMS map images offer a more effective representation of the NTFE. Specifically, GOMS map images naturally include spatial relationships between different elements in the network (\textit{e.g.,} road topology and bottleneck locations), which are essential for the extraction of rich spatial information in NTFE. In contrast, tabular data often struggles to capture these spatial relationships comprehensively and efficiently. The GOMS maps also contain extensive contextual information, such as geographic features and demographic patterns, that are difficult to express in tabular form. Typically, manual feature engineering is often required to extract the contextual information in tabular data, which can be both time-consuming and prone to human bias. Additionally, GOMS maps allow large amounts of data to be handled more efficiently. This scalability is particularly useful in NTFE, where traffic flow patterns need to be identified over large networks. More importantly, deep learning methods trained on map images tend to show satisfying generalization ability in various urban studies \citep{image_urban_embedding,image_accident,image_truck_estimation}, especially when faced with unseen network conditions. To summarize, these advantages make GOMS maps particularly powerful in NTFE, where the complexity and dynamic nature of flow patterns can be better captured and modeled.

In this study, we incorporate GOMS maps and observed traffic data into a deep learning method for NTFE. From the data perspective, we use three types of map images rather than traditional tabular data to incorporate richer and more comprehensive geographical and demographical information. These map images comprise OpenStreetMap (OSM), sensor distribution map, and population density map, including macro-level static geographical and demographical information such as POI, road topology, building footprints, sensor locations, population density, land cover, and so on. Furthermore, the observed traffic data are used to capture micro-level traffic dynamics. Importantly, the integration of these multi-source data constitutes the causes and consequences of traffic flow dynamics, which enables effective NTFE. From the methodological perspective, we develop an attention-based graph neural network with novel triple cross-attention and dense connection blocks to effectively fuse and extract the static geographical and demographical information from GOMS maps. Additionally, the observed traffic data are encoded through graph spatial and temporal blocks that embed the spatial-temporal traffic dynamics information. To validate the effectiveness of the proposed method, we test it across 15 cities in Europe and North America. Results show that the average, minimum, and maximum symmetric mean absolute percentage errors among all cities are 23\%, 17\%, and 27\%, respectively. The stable and satisfactory estimation accuracy in multiple cities demonstrates that the trade-off challenge can be successfully addressed using our approach.

To summarize, the major contributions of this paper are as follows:
\begin{itemize}
    \item This study represents the first attempt to leverage GOMS data for NTFE across cities, which not only effectively addresses the trade-off challenge but also lays the foundation for developing universal NTFE methods worldwide;
    
    \item Different from existing studies that rely on tabular data, we utilize map images that not only incorporate richer and more comprehensive geographical and demographical information but also give rise to a unified data format for NTFE;
    
    \item To effectively fuse spatial-temporal traffic dynamics with geographical and demographic information, a novel triple cross-attention block and dense connection block within an attention-based graph neural network is developed;
    
    \item We conduct a large-scale case study to test the accuracy and generality of the proposed method across 15 cities in Europe and North America. Results show satisfactory and stable estimation accuracy across these cities, which effectively addresses the trade-off challenge.
\end{itemize}

It is worth noting that NTFE based on GOMS maps represents the first attempt to acquire network-wide flow datasets across multiple cities, which provides valuable insights for transportation management in diverse urban environments. Specifically, these datasets offer an overview of traffic conditions and enable the identification of underlying traffic patterns. By understanding how traffic dynamics vary between cities, policymakers and planners can evaluate and compare transportation policies and infrastructure investments across different contexts. This facilitates the development of effective strategies tailored to each city, aimed at enhancing network connectivity, optimizing infrastructure development, and mitigating congestion. Furthermore, multi-city flow datasets promote information sharing among regional authorities, allowing cities to learn from each other's successes and challenges. This shared knowledge fosters collaborative planning efforts and coordinated transportation strategies for intercity transit systems. By leveraging collective insights, cities can better address common transportation issues and implement scalable solutions that benefit the broader metropolitan area. Importantly, GOMS maps serve as a critical foundation for developing generalized NTFE models in multiple cities. Their comprehensive and scalable representation of transportation activities enables more accurate and robust flow estimation across diverse urban environments. As more GOMS data become available, the performance and generalizability of NTFE are expected to improve significantly and warrant further investigation.

The rest of this paper is organized as follows. Section \ref{sec:liter} reviews the literature in NTFE. Section \ref{sec:problem} formulates the NTFE problem. Section \ref{sec:method} elaborates on the proposed attention-based graph neural network. Section \ref{sec:exp} focuses on experiments and evaluations of the proposed method. Finally, conclusions and future research directions are outlined in Section \ref{sec:conclusion}. 

\section{Literature review}\label{sec:liter}
Most of the existing studies on NTFE methods are validated in specific cities, the aforementioned trade-off challenge has not been well addressed. Specifically, early studies relied on the city-specific supplementary data sources for NTFE \citep{daganzo1997fundamentals, pendyala2000multi, lam2001activity, vuchic2007urban}. For example, \cite{pendyala2000multi} conducted multi-day and multi-period surveys to collect OD data, land-use patterns and household numbers for estimating the network-wide flow using the traditional four-step model. Despite their insights, these methods require significant effort and expense in data collection, and the reliance on strong model assumptions often fails to reflect real-world conditions \citep{google_flow,transfer_flow_3}, which limits their applications. With the development of sensor technologies, various ITS facilities have introduced numerous supplementary data, such as LPR data, video data and trajectory data, to tackle the NTFE problem \citep{cellphone_flow, machine_flow_1, camera_flow}. For example, \citet{camera_flow} employed convolutional neural networks (CNNs) to estimate the traffic flow based on the data collected from video surveillance cameras. However, these emerging data sources are only available in a few cities and therefore the associated methods suffer from the generality issue.

Recently, numerous studies have utilized FCD or observed traffic flow, available across multiple cities, for NTFE. Although these studies demonstrate the potential of applying NTFE across cities, their accuracy may be constrained due to the inherent limitations of the information provided by FCD or observed flow alone. For example, some studies formulated NTFE as a data imputation problem and solved the problem with various methods, such as autoregressive integrated moving average (ARIMA) \citep{arima_flow} and Principle Component Analysis (PCA) \citep{BPCA_flow, PPCA_flow, KPPCA_flow}. The spatial-temporal relationships of traffic data have also been exploited and constructed to enhance the NTFE through matrix/tensor decomposition \citep{tensor_decomposition_flow_2,tensor_decomposition_flow_3,tensor_decomposition_4_flow,tensor_decomposition_5_flow,tensor_decomposition_6_flow} and machine-learning-based methods \citep{machine_flow_1,machine_flow_2}. However, most of these methods relied on the historical records of traffic flow on roads, which may degrade the estimation accuracy for those unseen links in the training set. There are also studies that have estimated unobserved traffic flow by uncovering the latent relationship between traffic speed from FCD and observed flow. These efforts include not only estimating the link-based fundamental diagram \citep{ross1988traffic, kerner2009introduction, anuar2015estimating, google_flow} and MFD \citep{est_MFD_1,est_MFD_2,est_MFD_3} but also applying advanced deep learning techniques \citep{tgc_rnn_flow,sgmc_flow, GNN_new_flow}, such as GNNs, to exploit the spatiotemporal dependencies between flow and speed. Although neural networks can effectively capture these dependencies, their accuracy remains unsatisfactory when applied to unseen roads. Essentially, this is because the limited information provided by FCD alone is insufficient to comprehensively model traffic flow dynamics on unobserved roads.

There is also an emerging trend focusing on estimating traffic flow on unseen roads by leveraging multi-source data. The multi-source data includes but is not limited to geographical, demographical, sociological, and meteorological data, and these kinds of data can be acquired in various manners. For example, the global geographical data (\textit{e.g.,} road topology, POI, and land cover) can be obtained through several map providers such as OpenStreetMap, Google Maps, and Overture Maps Foundation. The local demographical and sociological data can be obtained through census \citep{demo_census} in several countries, while the global demographical data can be estimated through satellite images and statistical population model \citep{pop_map_get01,pop_map_get02}. The global meteorological data can be derived from the Global Historical Climatology Network hourly (GHCNh) dataset \citep{GHCNh}.
In the NTFE topic, \citet{multiple_regression_flow} proposed a multiple regression method based on network topology, Annual Average Daily Traffic (AADT), transport data, and FCD. The estimated travel time from Google Maps can also benefit traffic flow estimation \citep{google_flow}. \citet{open_data_flow_estimation} utilized traffic speed from Uber Movement and road static attributes from OSM for network flow inference. Several studies also have incorporated the POI, infrastructure, weather, road topology, and socioeconomic factors into the NTFE \citep{machine_flow_1, ssl_flow, data_driven_flow, search_flow_estimation}. For example, \citet{vision_flow} included visual information of probe vehicles for NTFE. \citet{flow_topo_pop} used topography information and population statistics. Other similar ideas of using geographical data to estimate traffic flow have also been implemented \citep{transfer_flow_1,gis_flow_1,gis_flow_2}. Nonetheless, most, if not all, of these studies validate their methods within specific cities because the adopted data are not unified and globally available. Consequently, they continue to face difficulties in overcoming the trade-off challenges.

\section{Problem statement} \label{sec:problem}
\begin{figure}[h]
    \centering
    \includegraphics[width=0.98\textwidth]{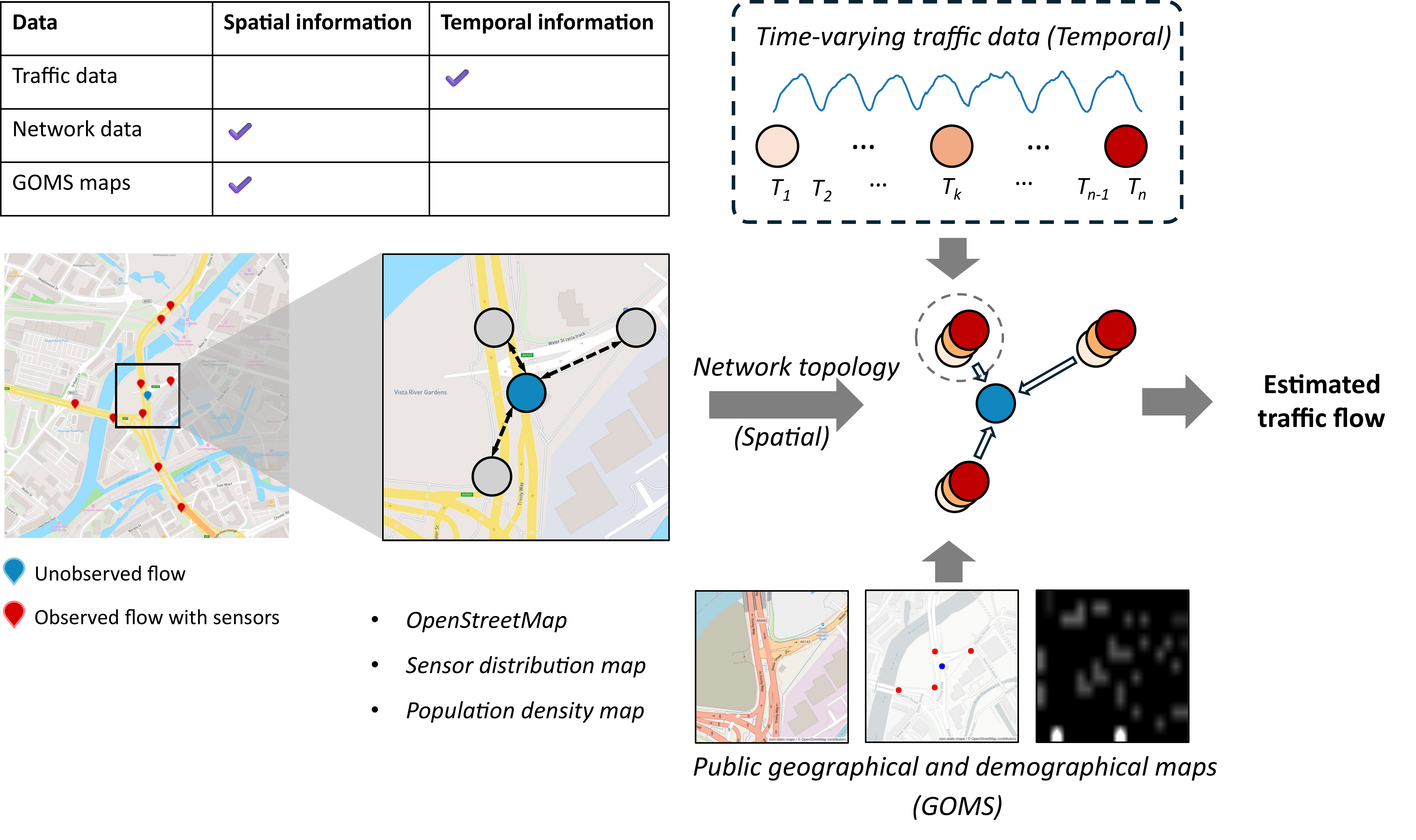}
    \caption{An example of the proposed NTFE framework.}
    \label{fig:problem_statement}
\end{figure}

In this section, we formulate the NTFE as a regression problem with multi-source data. The objective is to estimate the traffic flow on a road segment with: 1) The spatial and temporal traffic-related data on neighboring roads; and 2) The GOMS maps, including public geographical and demographical maps. An example of the proposed flow estimation framework is shown in Figure~\ref{fig:problem_statement}.
To illustrate the NTFE framework, we initially explain the multi-source data in Section~\ref{sec:data}, and then define the problem in Section~\ref{sec:problem_form}.

\subsection{Data description} \label{sec:data}
The utilized multi-source data are three-fold, public traffic data, network data, and GOMS maps, which are detailed as follows:


\paragraph{Traffic speed data}
In the existing literature, traffic speed data has been extensively utilized in NTFE due to the correlation between flow and speed, which can provide temporal information. Generally, the speed data can be obtained from FCD or stationary sensors from ITS across multiple cities, such as HKeMobility\footnote{\url{https://www.hkemobility.gov.hk/}} and PeMS\footnote{\url{https://dot.ca.gov/programs/traffic-operations/mpr/pems-source}}. The existing studies \citep{sgmc_flow,GNN_new_flow} also highlight that speed data can be sourced from third-party companies like Google and TomTom. The detailed description of the speed data in our work is provided in Section~\ref{sec:datasets}.

\paragraph{Network data}
The network data consists of a variety of spatial information about the road network that contributes to NTFE, which is well accepted in existing studies \citep{flow_topo_pop,gis_flow_1,gis_flow_2}. To ensure the network data is consistent across cities, we utilize only the fundamental network attributes, such as the number of lanes, and the longitude and latitude of roads. In this study, the network data are obtained from OSM and detailed in Section~\ref{sec:datasets}.

\paragraph{GOMS maps} 
The traditional geographical and demographical data contains plenty of structured and unstructured data which are laborious to process and standardized across to multiple cities. Furthermore, these data may not be available in one or more urban areas. To enhance the generality of NTFE methods across multiple cities, we incorporate three types of map images: OSM, sensor distribution map, and population density map. Our GOMS maps expand beyond traditional tabular data, offering richer and more comprehensive geographical and demographic information. Specifically, the OSM contains diverse land-use information such as road topology, POI, and land cover, which can provide general geographical information for NTFE. The sensor distribution map indicates the sensor locations and neighboring roads, providing an overview of the traffic monitoring setup for NTFE on a target road. The population density map draws the relative population density in a given area, which offers insights into human movements and potential transportation activities.

\subsection{Problem formulation} \label{sec:problem_form}
To better organize the input data with the consideration of spatial-temporal evolution, we first create a general spatial-temporal graph to model the relationship between the utilized multi-source data and network-wide flow. We define the spatial-temporal graph as $\mathcal{G} = (\mathcal{V}, \mathcal{E}, \mu_{a}, \mu_{i})$, where $\mathcal{V} = \left\{v_1, v_2, \cdots, v_{\mathcal{N}_v} \right\}$ is a set of nodes with size $\mathcal{N}_v$. $\mathcal{E} = \left\{\{ e_{1,t}, e_{2,t}, \cdots, e_{\mathcal{N}_e,t} \} \vert t = 1, 2, \cdots T \right\}$ is the set of time-dependent edges with given source nodes and target nodes. $\mathcal{N}_e$ and $T$ indicate the number of edges and study period, respectively. $\mu_{a}: \mathcal{V} \rightarrow \mathcal{A}_a$ is the node attribute function that maps the node set $\mathcal{V}$ into the node attribute set $\mathcal{A}_a \in \mathbb{R}^{\mathcal{N}_v \times \mathcal{N}_a \times T}$ (\textit{e.g.,} speed, lanes, longitude, latitude, \textit{etc}), where $\mathcal{N}_a$ is the number of attributes. $\mu_{i}: \mathcal{V} \rightarrow \mathcal{A}_i$ is the node image function that maps the node set $\mathcal{V}$ into the geographical and demographical images $\mathcal{A}_i \in \mathbb{R}^{\mathcal{N}_v \times H \times W \times (C_o + C_l + C_p)}$, where $H$ and $W$ are the height and width of images, and $C_o, C_l, C_p$ are the image channel of the OSM, sensor distribution map and population density map, respectively. For OSM and sensor distribution map, they are 3-channel RGB images, that is, $C_o = C_l = 3$. The population density map is a gray-scale image, that is, $C_p = 1$.

It is worth noting that the nodes and edges of $\mathcal{G}$ are not junctions and roads in the physical world. Actually, the nodes represent the fixed sensor locations, and the edges connect the nodes in the same local areas. Take the graph structure in the center of Figure~\ref{fig:problem_statement} as an example, a target node where traffic flow will be estimated is colored blue in the center of the image. The neighboring nodes are shown in grey and directly connected to the target node. However, they are not physically connected by roads or junctions in the real world.

Based on the above introduction, the proposed method for NTFE can be viewed as a function $\Phi: \mathcal{G}^{t-M, t} \rightarrow \mathcal{A}_{f}^{t}$, where $\mathcal{G}^{t-M, t}$ means a partial time-dependent graph from time $t-M$ to time $t$, and $\mathcal{A}_{f}^{t} \in \mathbb{R}^{\mathcal{N}_v}$ is the estimated traffic flow in the network at time $t$. Given a large study period $T$, it is expected that we can use the attention-based graph neural network to approximate $\Phi$. The estimation error at time $t$ is formulated as follows:
\begin{equation}
E_t = \left\Vert \Phi^{*}\left( \mathcal{G}^{t-M, t} \right)  - \hat{\mathcal{A}}_{f}^{t} \right\Vert, 
\end{equation}
where $\Phi^{*}$ represents the attention-based deep-learning approximator and $\hat{\mathcal{A}}_{f}^{t}$ denotes the observed ground truth of traffic flow in the network at time $t$.

\section{Method}\label{sec:method}
\begin{figure}[h]
    \centering
    \includegraphics[width=0.98\textwidth]{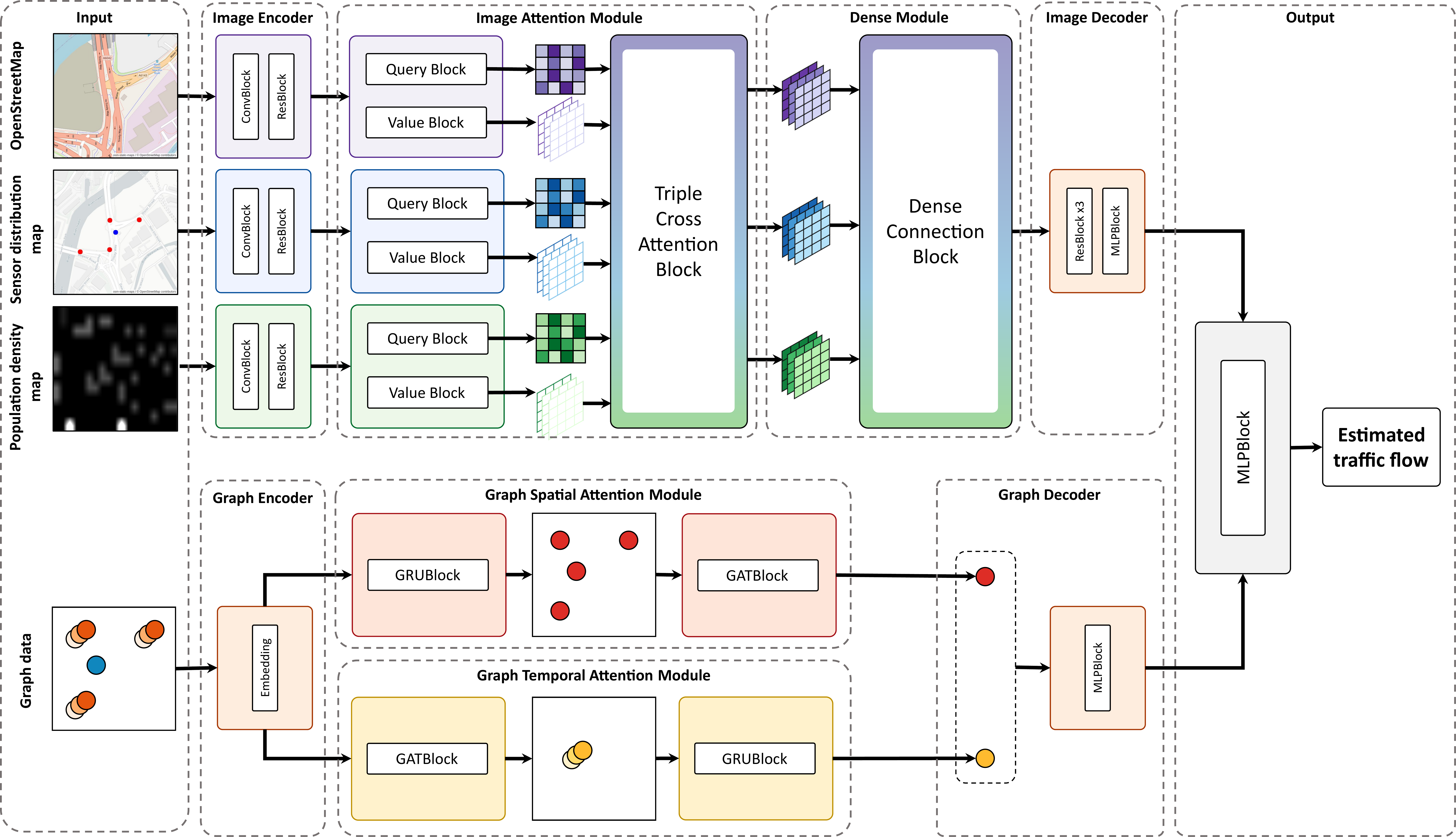}
    \caption{An overview of the proposed attention-based graph neural network with GOMS maps.}
    \label{fig:method}
\end{figure}

In this section, we develop an attention-based graph neural network for NTFE with GOMS maps. Specifically, the structure of the proposed method is shown in Figure~\ref{fig:method}. It can be seen that the neural network is mainly divided into two parts: the image processor and the graph processor. The image processor focuses on embedding map images for geographical and demographical information extraction. It consists of several modules: image encoder, image attention module, dense module, and image decoder. In the image attention module and dense module, two novel network blocks named Triple Cross-Attention Block (TCAB) and Dense Connection Block (DCB) are proposed to couple the geographical and demographical maps. The graph processor concentrates on embedding the time-dependent graph for NTFE. The graph encoder and decoder are developed to encode and decode the graph information. The Graph Spatial Attention Module (GSAM) and Graph Temporal Attention Module (GTAM) are proposed to further enhance the analysis of spatial-temporal correlations in graph data. Finally, the output from the image and graph processors are combined to provide the estimated network-wide traffic flow. As the network input contains various types of data, we extensively utilize the self- or cross-attention mechanism to fuse and condense information for NTFE. In the following sections, each module will be introduced in detail.

\subsection{Image processor}
In this section, four modules in the image processor named, image encoder, image attention module, dense module, and image decoder, will be elaborated separately.

\subsubsection{Image encoder}
The OSM, sensor distribution map, and population density map are pre-encoded by three individual image encoders. We employ a convolution block (ConvBlock) and a residual block (ResBlock) in work \citep{resnet} to embed these images, shown in Equation~\ref{eq:image_encoder}:
\begin{equation}
    \begin{aligned}
        I^{o,e}_{i} &= \mathcal{F}_{r}^{o} \left ( \mathcal{F}_{c}^{o} \left( I^{o,r}_{i} \right) \right), \\
        I^{l,e}_{i} &= \mathcal{F}_{r}^{l} \left ( \mathcal{F}_{c}^{l} \left( I^{l,r}_{i} \right) \right), \\
        I^{p,e}_{i} &= \mathcal{F}_{r}^{p} \left ( \mathcal{F}_{c}^{p} \left( I^{p,r}_{i} \right) \right), \\
    \end{aligned}
    \label{eq:image_encoder}
\end{equation}
where $I^{o,r}_{i}, I^{l,r}_{i}, I^{p,r}_{i}$ are the $i$th OSM, sensor distribution map, and population density map images, respectively. $I^{o,e}_{i}, I^{l,e}_{i}, I^{p,e}_{i}$ are the corresponding encoded map features. $\mathcal{F}_{c}^{o}, \mathcal{F}_{c}^{l}, \mathcal{F}_{c}^{p}$ and $\mathcal{F}_{r}^{o}, \mathcal{F}_{r}^{l}, \mathcal{F}_{r}^{p}$ are the ConvBlocks and ResBlocks, respectively, for the associated OSM, sensor distribution map and population density map. Please refer to the work by \citet{resnet} for the detailed settings of the ConvBlocks and ResBlocks.

\subsubsection{Image attention module}
There are two aims for the image attention module: 1) The image attention module focuses on extracting geographical and demographical information within long-range and multi-level map images; 2) The image attention module unifies the geographical and demographical information across images. In this block, the encoded images are embedded with query blocks and value blocks simultaneously. The output feature maps from these blocks are then processed by the TCAB for further extraction and combination.

The query block delivers an attention score map $I_{i}^{m, q} \in \mathbb{R}^{H \times W \times C_m}$ which indicates the extent to which the block focuses on the $m$th image on pixel $(j,k)$ when synthesizing the channel $c$, $1 \leq j \leq H, 1 \leq k \leq W, 1 \leq c \leq C_m$. The formulation of the query block is shown in Equation~\ref{eq:query_block}:
\begin{equation}
    \begin{aligned}
        I^{m,q}_{i} &= \text{softmax}\left( \mathcal{F}_{q}^{m} \left( I^{m,e}_{i} \right)^T \otimes \mathcal{F}_{k}^{m} \left( I^{m,e}_{i} \right) \right), \quad \forall m \in \left\{ o, l, p \right\}, \\
        \text{softmax}(I^{m}_{j,k,c}) &= \frac{\exp \left( I_{j,k,c}^{m} \right)}{\sum_{j=1}^{H} \sum_{k=1}^{W} \exp \left( I_{j,k,c}^{m} \right)}, \quad  \forall 1 \leq j \leq H, 1 \leq k \leq W, 1 \leq c \leq C_m, m \in \left\{ o, l, p \right\},
    \end{aligned}
    \label{eq:query_block}
\end{equation}
where $\mathcal{F}_{q}^{m}$ and $\mathcal{F}_{k}^{m}$ are two $1\times1$ convolution blocks for different feature maps ($o$ for OSM, $l$ for sensor distribution map, and $p$ for population density map). $\otimes$ denotes the tensor multiplication and $\text{softmax()}$ is a softmax function. All elements in $I^{m,q}_{i}$ should range from 0 to 1. The value block further encodes the feature maps, shown in Equation~\ref{eq:value_block}:
\begin{equation}
    I^{m,v}_{i} = \mathcal{F}_{v}^{m} \left( I^{m,e}_{i} \right), \forall m \in \left\{ o, l, p \right\},
    \label{eq:value_block}
\end{equation}
where $\mathcal{F}_{v}^{m}$ is also a $1\times1$ convolution block for different feature maps.

\begin{figure}[h]
    \centering
    \includegraphics[width=0.95\textwidth]{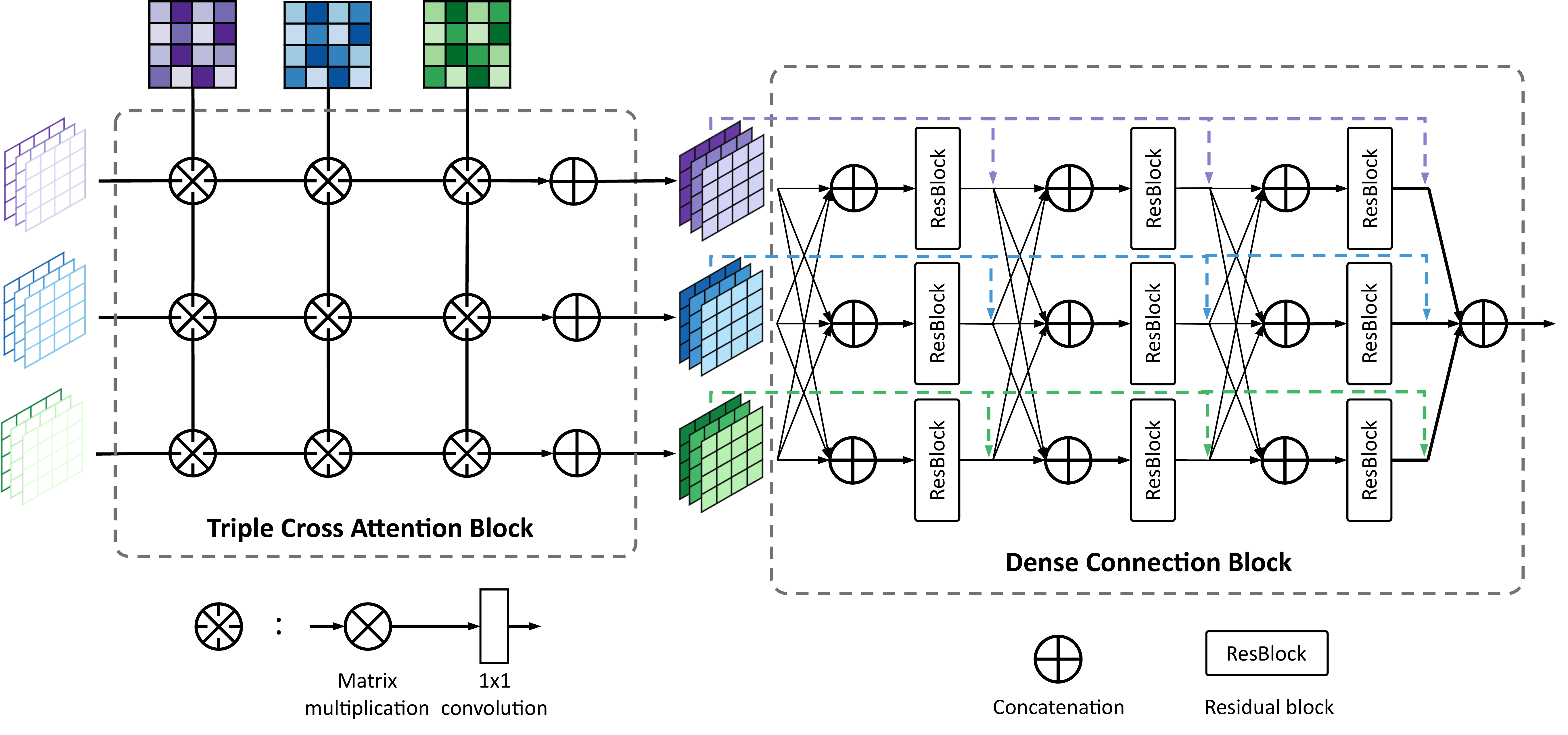}
    \caption{The pipeline of the TCAB and the DCB.}
    \label{fig:TCAB_DCB}
\end{figure}

While the query block and value block concentrate on extracting geographical and demographical information within long-range and multi-level, the TCAB focuses on unifying the information from different feature maps. The pipeline of the TCAB is on the left-hand side of Figure~\ref{fig:TCAB_DCB}. 
In TCAB, each feature map from the value blocks ($I^{o,v}_{i}$, $I^{l,v}_{i}$ and $I^{p,v}_{i}$) will react with feature maps from the query blocks ($I^{o,q}_{i}$, $I^{l,q}_{i}$ and $I^{p,q}_{i}$). 

The formulation of the TCAB is shown in Equations ~\ref{eq:TCAB_1} and ~\ref{eq:TCAB_2}:
\begin{equation}
    \begin{aligned}
        \mathcal{I}_{i}^{t} &= \left[ I^{o,q}_{i}, I^{l,q}_{i}, I^{p,q}_{i} \right]^T \textcircled{\textreferencemark} \left[ I^{o,v}_{i}, I^{l,v}_{i}, I^{p,v}_{i} \right]\\
        &= \begin{bmatrix}
            \mathcal{F}_{t}^{o,o}\left( I^{o,q}_{i} I^{o,v}_{i} \right) & \mathcal{F}_{t}^{o,l}\left( I^{o,q}_{i} I^{l,v}_{i} \right) & \mathcal{F}_{t}^{o,p}\left( I^{o,q}_{i} I^{p,v}_{i} \right) \\
            \mathcal{F}_{t}^{l,o}\left( I^{l,q}_{i} I^{o,v}_{i} \right) & \mathcal{F}_{t}^{l,l}\left( I^{l,q}_{i} I^{l,v}_{i} \right) & \mathcal{F}_{t}^{l,p}\left( I^{l,q}_{i} I^{p,v}_{i} \right) \\
            \mathcal{F}_{t}^{p,o}\left( I^{p,q}_{i} I^{o,v}_{i} \right) & \mathcal{F}_{t}^{p,l}\left( I^{p,q}_{i} I^{l,v}_{i} \right) & \mathcal{F}_{t}^{p,p}\left( I^{p,q}_{i} I^{p,v}_{i} \right)
        \end{bmatrix},
    \end{aligned}
    \label{eq:TCAB_1}
\end{equation}
\begin{equation}
    \begin{aligned}
        I_{i}^{o,t} &= \text{concat} \left( 
        \mathcal{F}_{t}^{o,o}\left( I^{o,q}_{i} I^{o,v}_{i} \right), \mathcal{F}_{t}^{l,o}\left( I^{l,q}_{i} I^{o,v}_{i} \right), 
        \mathcal{F}_{t}^{p,o}\left( I^{p,q}_{i} I^{o,v}_{i} \right)
        \right), \\
        I_{i}^{l,t} &= \text{concat} \left( 
        \mathcal{F}_{t}^{o,l}\left( I^{o,q}_{i} I^{l,v}_{i} \right),
        \mathcal{F}_{t}^{l,l}\left( I^{l,q}_{i} I^{l,v}_{i} \right),
        \mathcal{F}_{t}^{p,l}\left( I^{p,q}_{i} I^{l,v}_{i} \right)
        \right), \\
        I_{i}^{p,t} &= \text{concat} \left( 
        \mathcal{F}_{t}^{o,p}\left( I^{o,q}_{i} I^{p,v}_{i} \right),
        \mathcal{F}_{t}^{l,p}\left( I^{l,q}_{i} I^{p,v}_{i} \right),
        \mathcal{F}_{t}^{p,p}\left( I^{p,q}_{i} I^{p,v}_{i} \right)
        \right), \\
    \end{aligned}
    \label{eq:TCAB_2}
\end{equation}
where $\mathcal{F}_{t}^{a,b}$ denotes a calculation unit including the tensor multiplication and $1\times1$ convolution. The $a$ and $b$ indicate the image type which is either the OSM, sensor distribution map or population density map. The $\text{concat}$ function represents the concatenation process of feature maps. $I_{i}^{o,t}, I_{i}^{l,t}, I_{i}^{p,t}$ are the feature maps of the OSM, sensor distribution map, and population map after the TCAB, respectively.

\subsubsection{Dense module}
To exhaustively use the fruitful geographical and demographical information, we develop a DCB, including multiple add and concatenation operations. The DCB contains sequential operations with similar structures. The formulation of the $j$th layer in DCB is shown in Equation~\ref{eq:DCB}:
\begin{equation}
    \begin{aligned}
        I_{i,j+1}^{m,d} &= \mathcal{F}_{j}^{m,d} \left(\text{concat}\left( I_{i,j}^{o,d}, I_{i,j}^{l,d}, I_{i,j}^{p,d} \right) \right) + \mathcal{F}_{j}^{m,s} \left(I_{i,j}^{m,d} \right), \quad \forall j = 1, \dots, \mathcal{N}_d, m \in \left\{ o, l, p \right\}, \\
        I_{i,0}^{o,d} &= I_{i}^{o,t}; I_{i,0}^{l,d} = I_{i}^{l,t}; I_{i,0}^{p,d} = I_{i}^{p,t}, \\
        I_{i}^{d} &= \text{concat}\left( I_{i,\mathcal{N}_d}^{o,d}, I_{i,\mathcal{N}_d}^{l,d}, I_{i,\mathcal{N}_d}^{p,d} \right),
    \end{aligned}
    \label{eq:DCB}
\end{equation}
where $I_{i,j}^{o,d}, I_{i,j}^{l,d}, I_{i,j}^{p,d}$ are the input feature maps of the OSM, sensor distribution map, and population density map into the $j$th layer. Here, we define the $0$th input feature as the output from the TCAB. $\mathcal{N}_d$ is the layer number of the DCB.  $I_{i,j+1}^{m,d}$ is the expected output feature maps after the $j$th operation. $\mathcal{F}_{j}^{m,d}$ is a ConvBlock for the concatenated feature maps, and $\mathcal{F}_{j}^{m,s}$ is a down-sampling layer consisting of a convolution block and max-pooling layer. The final output feature map from the DC that contains condensed geographical and demographical information $I_{i}^{d}$ is formulated as the concatenations of the features maps from the last operation.

\subsubsection{Image decoder}
The image decoder further compresses the feature maps (\textit{i.e.,} tensors) into embeddings (\textit{i.e.,} vectors) for calculating the traffic flow with graph embeddings. The formulation of the image decoder is shown in Equation~\ref{eq:image_decoder}:
\begin{equation}
    I_{i}^{p} = \mathcal{F}^{p}\left( I_{i}^{d} \right),
    \label{eq:image_decoder}
\end{equation}
where $\mathcal{F}^{p}$ is the image decoder consisting of a series of ResBlocks and Multilayer perception (MLPBlock), and $I_{i}^{p}$ is the final embedding for geographical and demographical maps.

\subsection{Graph processors}
In this section, four modules in the graph processor named, graph encoder, GSAM, GTAM, and graph decoder, will be elaborated separately.

\subsubsection{Graph encoder}
The graph encoder embeds the attributes of nodes in $\mathcal{G}$  into a higher dimension, which is formulated in Equation~\ref{eq:node_encoder}:
\begin{equation}
    V_{i, t}^{e} = \mathcal{F}^{e} \left( V_{i, t}^{r} \right),
    \label{eq:node_encoder}
\end{equation}
where $V_{i,t}^{r} \in \mathcal{A}_{a}$ is the initial $i$th node attribute at time $t$, $\mathcal{F}^{e}$ is the graph encoder and $V_{i,t}^{e}$ is the encoded node embedding.

\subsubsection{Graph Spatial Attention Module}
The GSAM extracts the spatial information of a target node. The node embeddings in a time interval are initially aggregated through a Gate Recurrent Unit (GRU) layer, and the aggregated node embeddings are further fed into the Graph Attention Network (GAT) layer to extract the spatial correlation between adjacent nodes. The formulation of GSAM is shown in Equation~\ref{eq:GSAM}:
\begin{equation}
    V_{i, t}^{s} = \mathcal{F}^{s,s} \left( \mathcal{F}^{s,t} \left( V_{i, t-M, t}^{e} \right) \right), \quad \forall t = M+1, \dots, T,
    \label{eq:GSAM}
\end{equation}
where $V_{i, t-M, t}^{e}$ denotes the embedding of node $i$ from time $t-M$ to time $t$. $\mathcal{F}^{s,s}$ and $\mathcal{F}^{s,t}$ are the GAT and GRU layers separately. $V_{i, t}^{s}$ is the spatial node embedding.

\subsubsection{Graph Temporal Attention Module}
The GTAM mainly focuses on extracting the temporal information for a target node. The node embeddings in a time interval are initially encoded spatially through GAT, and the GRU layer is included to aggregate the node temporal embeddings consequently. The formulation of the GTAM is shown in Equation~\ref{eq:GTAM}:
\begin{equation}
    V_{i, t}^{t} = \mathcal{F}^{t,t} \left( \mathcal{F}^{t,s} \left( V_{i, t-M, t}^{e} \right) \right), \quad \forall t = M+1, \dots, T
    \label{eq:GTAM}
\end{equation}
where $\mathcal{F}^{t,t}$ and $\mathcal{F}^{t,s}$ are the GRU and GAT layers. $\mathcal{V}_{i,t}^t$ is the temporal node embedding.

\subsubsection{Graph decoder}
The graph decoder fuses the spatial and temporal embeddings of the target with an MLPBlock, which is shown in Equation~\ref{eq:graph_decoder}:
\begin{equation}
    V_{i,t}^{d} = \mathcal{F}_{g} \left(\text{concat} \left( V_{i, t}^{s}, V_{i, t}^{t}\right) \right),
    \label{eq:graph_decoder}
\end{equation}
where $\mathcal{F}_{g}$ is the graph decoder and $V_{i,t}^{d}$ is the merged node embedding.

\subsection{Output}
The final block outputs the estimated traffic flow given the node embeddings and image embeddings through an MLPBlock, which is shown in Equation~\ref{eq:output}:
\begin{equation}
    y_{i,t} = \mathcal{F}_{o} \left( \text{concat} \left( V_{i, t}^{d}, I_{i}^{p}\right) \right),
    \label{eq:output}
\end{equation}
where $\mathcal{F}_{o}$ is the output MLPBlock and $y_{i,t}$ is the estimated traffic flow of node $i$ at time $t$.

Overall, we aim to fully extract and condense valid information for traffic flow estimation with multiple input data using the proposed method. For the image part, we design the TCAB to extract the multilevel information and fuse it through the DCB. For the graph data part, we utilize the GSAM and GTAM to encode the spatial-temporal features. The final output block will decode all the information to the network flow.

\section{Experiments}\label{sec:exp}
In this section, we evaluate the proposed method across 15 cities in Europe and North America. In Section \ref{sec:datasets}, we introduce the utilized data sources. In Section~\ref{sec:baseline_methods}, we present the baseline methods for comparison with the proposed method. Section~\ref{sec:exp_setting} elaborates on the details of the experimental settings. The experimental results are summarized in Section~\ref{sec:exp_result}. Importantly, the ablation study implemented in Section~\ref{sec:abl_study} demonstrates that the utilization of GOMS maps is necessary for accurate NTFE. A sensitivity analysis is conducted in Section~\ref{sec:sen_analysis}. 

\subsection{Datasets} \label{sec:datasets}
In this study, we use the datasets of 15 cities in Europe and North America to conduct the experiment. Specifically, there are 6 cities in Europe which are, Birmingham, Bolton, Essen, Innsbruck, Manchester, and Rotterdam, and 9 cities in North America, namely Fresno, Los Angeles, Oakland, Riverside, Sacramento, Salinas, San Diego, San Jose, and Stockton. The OSM, sensor distribution map, and population density map are introduced as follows.

\subsubsection{Traffic data and network data}
\begin{figure}
    \centering
    \includegraphics[width=1\textwidth]{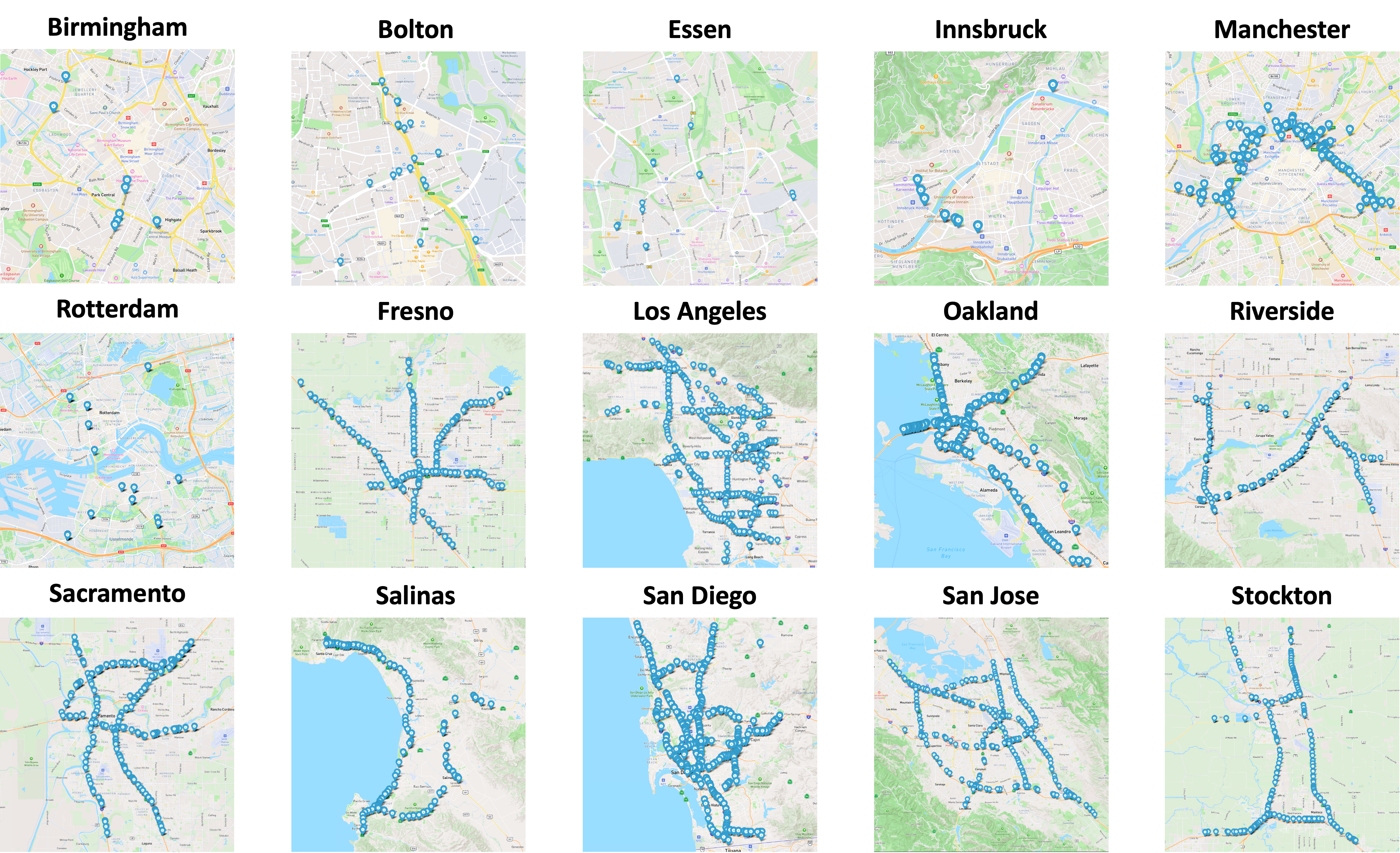}

    \caption{Snapshots of road segments with the placement of fixed sensors.}
    \label{fig:cities}
\end{figure}

The public traffic data and network data are derived from two widely accepted datasets. The data in Europe is from the UTD19 traffic dataset \citep{utd19}, and the data in North America is from the Performance Measurement System (PeMS) in the California Department of Transportation (Caltrans). 
\begin{itemize}
    \item \texttt{UTD19}: The UTD19 dataset contains traffic-related data, including speed, flow and network attributes from more than 40 cities worldwide. The data ranges from 2017 to 2019. The time interval for data collection ranges from 3 to 5 minutes.
    \item \texttt{PeMS}: The PeMS dataset includes traffic-related data including speed, flow and network attributes (the same as the UTD19) from 12 districts in California, US. The data used in this study ranges from July to August, 2021. The time interval for data collection is 5 minutes.
\end{itemize}
Table~\ref{tab:datasets} summarizes the number of road segments used for training and testing in each city, and
\begin{table}[h]
    \centering
    \caption{The number of road segments in cities.}
    \begin{tabular}{p{0.12\textwidth}|p{0.12\textwidth}p{0.20\textwidth}p{0.20\textwidth}p{0.20\textwidth}}
    \hline
        City & Country & Total road segments & Roads for training & Roads for testing \\
        \hline
        Birmingham & UK & 8 & 6 & 2 \\
        Bolton & UK & 19 & 14 & 5 \\
        Essen & Germany & 14 & 11 & 3 \\
        Innsbruck & Austria & 9 & 7 & 2 \\
        Manchester & UK & 122 & 100 & 22 \\
        Rotterdam & Netherlands & 63 & 51 & 12 \\
        Fresno & US & 217 & 175 & 42 \\
        Los Angeles & US & 422 & 347 & 75 \\
        Oakland & US & 204 & 154 & 50 \\
        Riverside & US & 375 & 323 & 52 \\
        Sacramento & US & 375 & 315 & 60 \\
        Salinas & US & 193 & 163 & 30 \\
        San Diego & US & 429 & 368 & 61 \\
        San Jose & US & 417 & 357 & 60 \\
        Stockton & US & 259 & 215 & 44 \\
        \hline
        Sum & - & 3,126 & 2,606 & 520 \\
    \hline
    \end{tabular}
    \label{tab:datasets}
\end{table}
Figure~\ref{fig:cities} visualizes the road segments with annotated sensor data (\textit{i.e.,} traffic flow) in each city.
Note that we have all ground true flow on all road segments in both training and testing set in all cities for evaluation purposes. Though the fixed sensors do not cover the entire road network in most cities in Europe, and several cities in North America, we can still evaluate the performance of the proposed method under different sensor coverage. Furthermore, we can supplement the traffic flow on unobserved roads, which is the major motivation of this paper.

\subsubsection{GOMS maps}
\begin{figure}[h]
  \centering    
  \subfigure[] {
   \label{fig:poi_15}     
  \includegraphics[width=0.3\columnwidth]{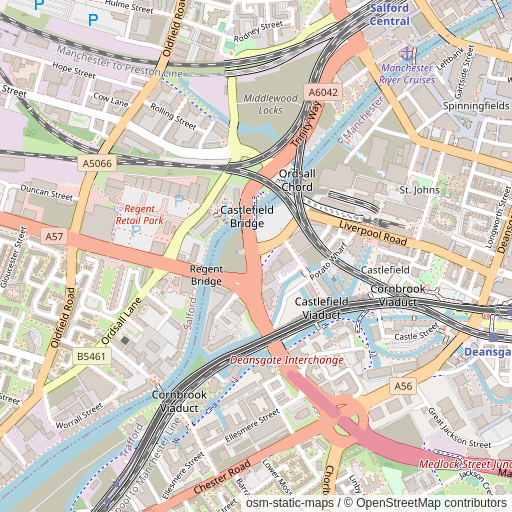}
  }
  \subfigure[] { 
  \label{fig:poi_16}     
  \includegraphics[width=0.3\columnwidth]{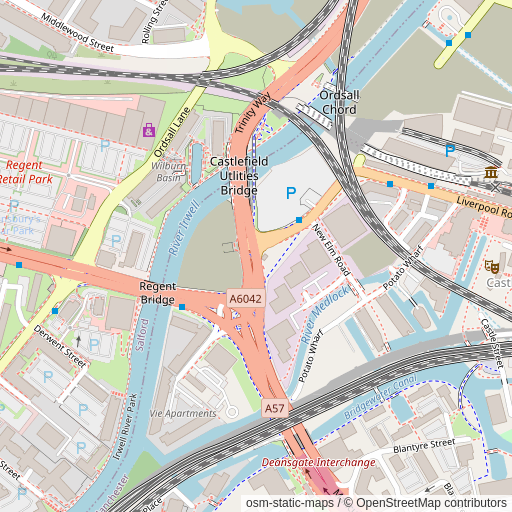}
  }
  \subfigure[] { 
  \label{fig:poi_17}     
  \includegraphics[width=0.3\columnwidth]{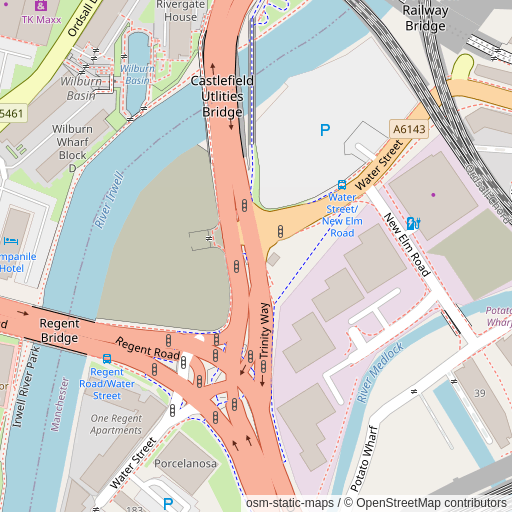}
  }
  \subfigure[] {
   \label{fig:point_15}     
  \includegraphics[width=0.3\columnwidth]{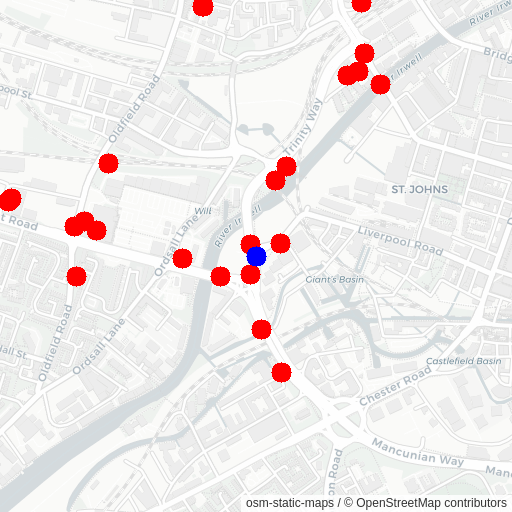}
  }
  \subfigure[] { 
  \label{fig:point_16}     
  \includegraphics[width=0.3\columnwidth]{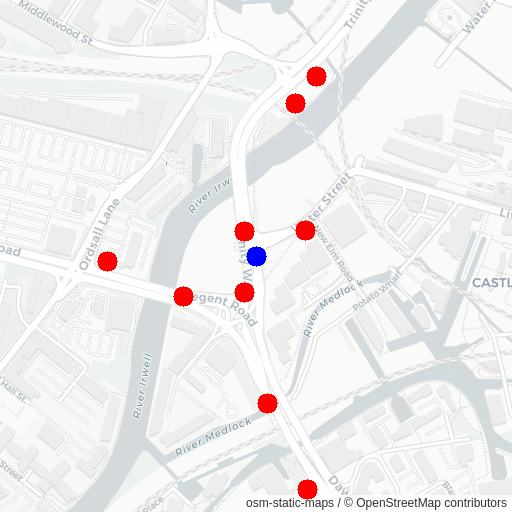}
  }
  \subfigure[] { 
  \label{fig:point_17}     
  \includegraphics[width=0.3\columnwidth]{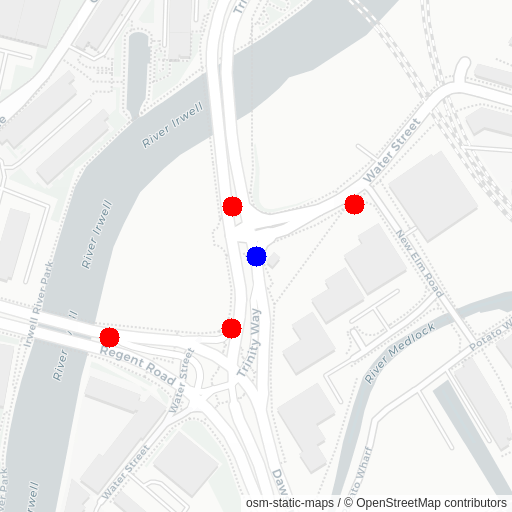}
  }
  \subfigure[] {
   \label{fig:pop_15}     
  \includegraphics[width=0.3\columnwidth]{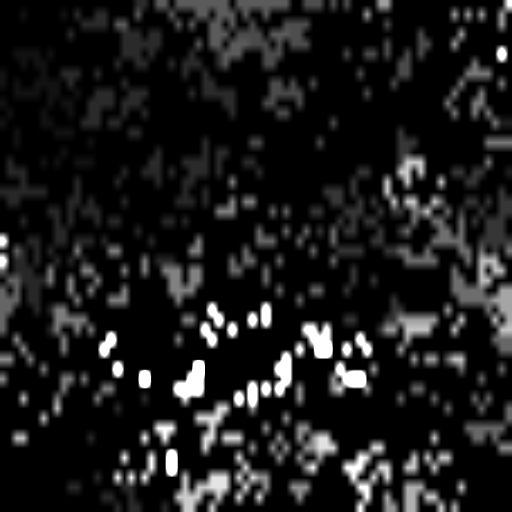}
  }
  \subfigure[] { 
  \label{fig:pop_16}     
  \includegraphics[width=0.3\columnwidth]{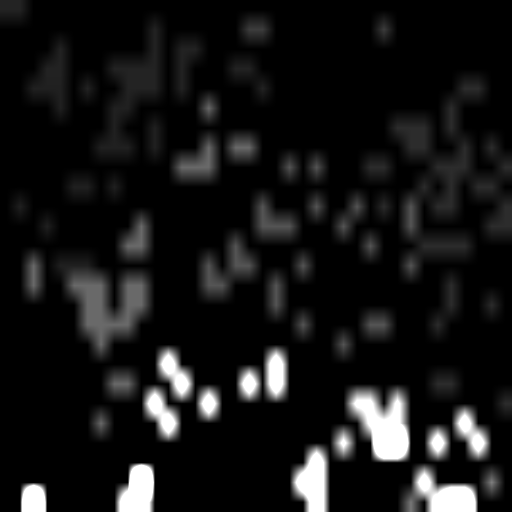}
  }
  \subfigure[] { 
  \label{fig:pop_17}     
  \includegraphics[width=0.3\columnwidth]{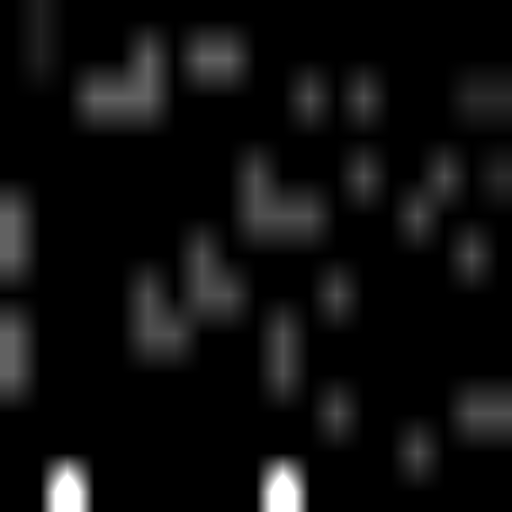}
  }
  \label{fig:image_example}
  \caption{Snapshots from GOMS maps in Manchester. \ref{fig:poi_15}-\ref{fig:poi_17} are OSMs with zoom level of 15-17. \ref{fig:point_15}-\ref{fig:point_17} are sensor distribution maps with zoom levels of 15-17.\ref{fig:pop_15}-\ref{fig:pop_17} are population density maps with zoom levels of 15-17.}
\end{figure}

\begin{itemize}
    \item \texttt{OpenStreetMap}: The OSM is a free and publicly available map consisting of various geographical information such as land cover, road topology, POI, \textit{etc}. In this experiment, the OSM images are snapshotted using the ``osm-static-map''\footnote{\url{https://github.com/jperelli/osm-static-maps}} tool with different scales and resolutions shown in Figures~\ref{fig:poi_15}-\ref{fig:poi_17}; 
    \item \texttt{Sensor distribution map}: The sensor distribution map consists of locations of neighboring sensors and background maps of the target road shown in Figures~\ref{fig:point_15}-\ref{fig:point_17}. The sensor locations colored with red are drawn using the longitude and latitude information in the UTD19 and PeMS datasets. The background maps are also derived from the OSM using the ``osm-static-map'' tool. In contrast to the original OSM images, the background maps are sourced from an alternative raster tile provider with lighter colors to better highlight the information pertaining to the sensor locations.
    \item \texttt{Population density map}: The population density maps, shown in Figures~\ref{fig:pop_15}-\ref{fig:pop_17}, are obtained from Meta\textsuperscript{\textcopyright} (previously Facebook) based on satellite images, which can be globally and openly downloaded all over the world. Moreover, it is worth noting that the OSM and sensor distribution map are 3-channel RGB images, while the density maps are 1-channel gray-scale images. The gray-scale value indicates the population in a local area, a factor widely recognized as one of the primary determinants of transportation activities. The pixel size of the density map corresponds to 30 square meters in the real world, and thus the value of each pixel indicates the population per 30 square meters.
\end{itemize}

\subsection{Baseline methods} \label{sec:baseline_methods}
In this section, we present the baseline methods for comparison with our method.
\begin{itemize}
    \item Spatial Average (SA): The SA estimates the unobserved road segments based on the average of neighboring observed road segments.
    \item Long short-term memory (LSTM): The LSTM network is a common deep-learning method for time-series forecast and estimation, which can achieve decent performance \citep{LSTM}. However, this method can hardly encode the spatial correlation of data in the graph. The node data in the graph are dismantled as individual data records, which can be batched for training.
    \item Graph Convolution Matrix Completion (GCMC): The GCMC method was initially proposed for the recommending system \citep{GCMC}. It can also be used for NTFE. It advances the traditional matrix completion methods by incorporating spatial information and graph convolution. To achieve this, the origin graph is transferred to a bipartite graph. There are two sets of nodes, road ID and time, where the links between these two node sets are the traffic flow. Hence, the goal changes to estimate the link value, which is equivalent to estimating the traffic flow in the origin graph.
    \item Graph Attention Network version 2 (GATv2): The GAT succeeds in representation learning with graphs by incorporating the attention mechanism. The GATv2 \citep{GATv2} is superior to GATv1 \citep{GATv1} by incorporating the dynamic graph attention mechanism rather than the static one. 
\end{itemize}

\subsection{Experiment settings} \label{sec:exp_setting}
All experiments are conducted on a desktop with Intel Core I9-13900K CPU @5.4GHz $\times$ 8, 4.3GHz $\times$ 16, 4800MHz $\times$ 2 $\times$ 32GB RAM, GeForce RTX 3090 $\times$ 2, 1TB SSD. The roads in all the cities are divided into training roads and testing roads, which are shown in Table~\ref{tab:datasets}. In the training stage, the data from training roads are constructed into a graph, while the data in the testing roads are ignored. In the testing stage, we randomly selected 20\% of the whole time intervals as the testing period. The time interval between each data point is set to 5 minutes. One-hour historical graph data is selected to estimate the current traffic flow, meaning that the time window $M = 12$.

The objective of the proposed method is to minimize the Mean Square Error (MSE) (defined in Equation~\ref{eq:criteria}) between the estimated traffic flow $y_{*,t}$ and the ground truth $\hat{y}_{*,t}$. To optimize the proposed method, we leverage the Adam optimizer \citep{adam} with a learning rate of $1 \times 10^{-4}$. Other hyperparameters in the Adam optimizer are referred to as the default settings. The embedding size of the graph data and images after the image and graph decoder is set to 512. We select three criteria to evaluate the experimental results, which are Root Mean Square Error (RMSE), Mean Absolute Error (MAE), and Symmetric Mean Absolute Percentage Error (SMAPE), which are formulated in following equations. The units of the RMSE and MAE are the vehicle per hour per lane (veh/hour/lane).
\begin{align}
        \text{MSE} \left(y_{*,t}, \hat{y}_{*,t} \right) &= \frac{1}{\mathcal{N}_v} \sum_{i=1}^{\mathcal{N}_v} \left( y_{*,t} - \hat{y}_{i,t} \right)^2; \\
        \text{RMSE} \left(y_{*,t}, \hat{y}_{*,t} \right) &= \sqrt{\frac{1}{\mathcal{N}_v} \sum_{i=1}^{\mathcal{N}_v} \left( y_{*,t} - \hat{y}_{*,t} \right)^2}; \\
        \text{MAE} \left(y_{*,t}, \hat{y}_{*,t} \right) &= \frac{1}{\mathcal{N}_v} \sum_{i=1}^{\mathcal{N}_v} \left\vert y_{*,t} - \hat{y}_{*,t} \right\vert; \\
        \text{SMAPE} \left(y_{*,t}, \hat{y}_{*,t} \right) &= \frac {100\%}{\mathcal{N}_v} \sum_{i=1}^{\mathcal{N}_v} \frac{\left\vert y_{*,t} - \hat{y}_{*,t} \right\vert}{\left( y_{*,t} + \hat{y}_{*,t}\right) / 2}.
    \label{eq:criteria}
\end{align}

\subsection{Experimental results} \label{sec:exp_result}
\begin{figure}
    \centering
    \includegraphics[width=1\textwidth]{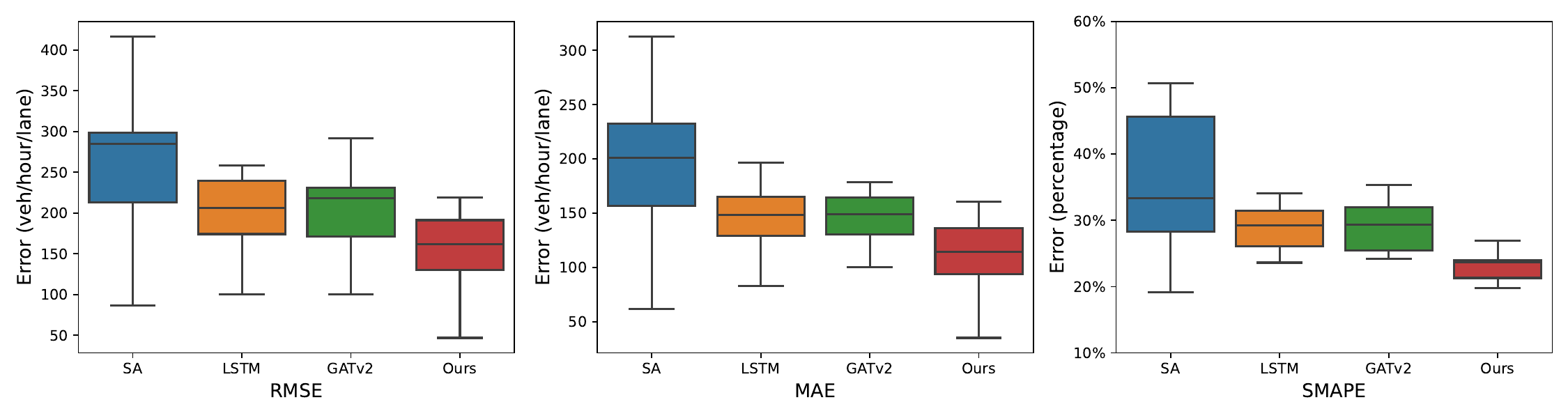}
    \caption{Boxplots of the estimation errors in all cities. The units for RMSE and MAE are veh/hour/lane.}
    \label{fig:benchmark_summary}
\end{figure}

To evaluate the estimation performance of the proposed method, we initially compare the results from the proposed method with those from the aforementioned baseline methods. The comparisons within cities of Europe and North America are shown in Tables~\ref{tab:benchmark_eu} and ~\ref{tab:benchmark_na}, respectively. In these tables, the RMSE, MAE and SMAPE of each method are listed from top to bottom, and the units for RMSE and MAE are vehicles per hour per lane. It can be seen that our method can achieve the minimal RMSE, MAE and SMAPE in all cities in Europe and North America. Moreover, all SMAPE of our method is below or around 25\% except Rotterdam, Riverside and Salinas. We also display the boxplots of estimation errors for the different methods in Figure~\ref{fig:benchmark_summary}. Here, we do not include the results using the GCMC since the error is much larger than other methods. Although the GCMC should be an efficient and accurate method to impute the missing values in traffic sensor data, it fails to generalize the data from unseen sensors in the training set.

In the evaluation, the average RMSE and MAE are $154.09$ and $111.03$ veh/hour/lane and the SMAPE is $22.81\%$. Our method performs the best in Essen. The RMSE and MAE are $46.77$ and $35.13$ veh/hour/lane and the SMAPE is $16.79\%$. The maximal error is acquired in Salinas, where the RMSE and MAE are $140.42$ and $97.63$ veh/hour/lane and the SMAPE is $26.93\%$. The maximal error using the proposed method is still comparable to the median error using the benchmark methods. Moreover, the gap between the maximal and minimal SMAPEs using the proposed method is smaller than the ones using benchmark methods, demonstrating that our method succeeds in both accuracy and generality across multiple cities.

\begin{figure}[ht]
    \centering
    \includegraphics[width=0.95\textwidth]{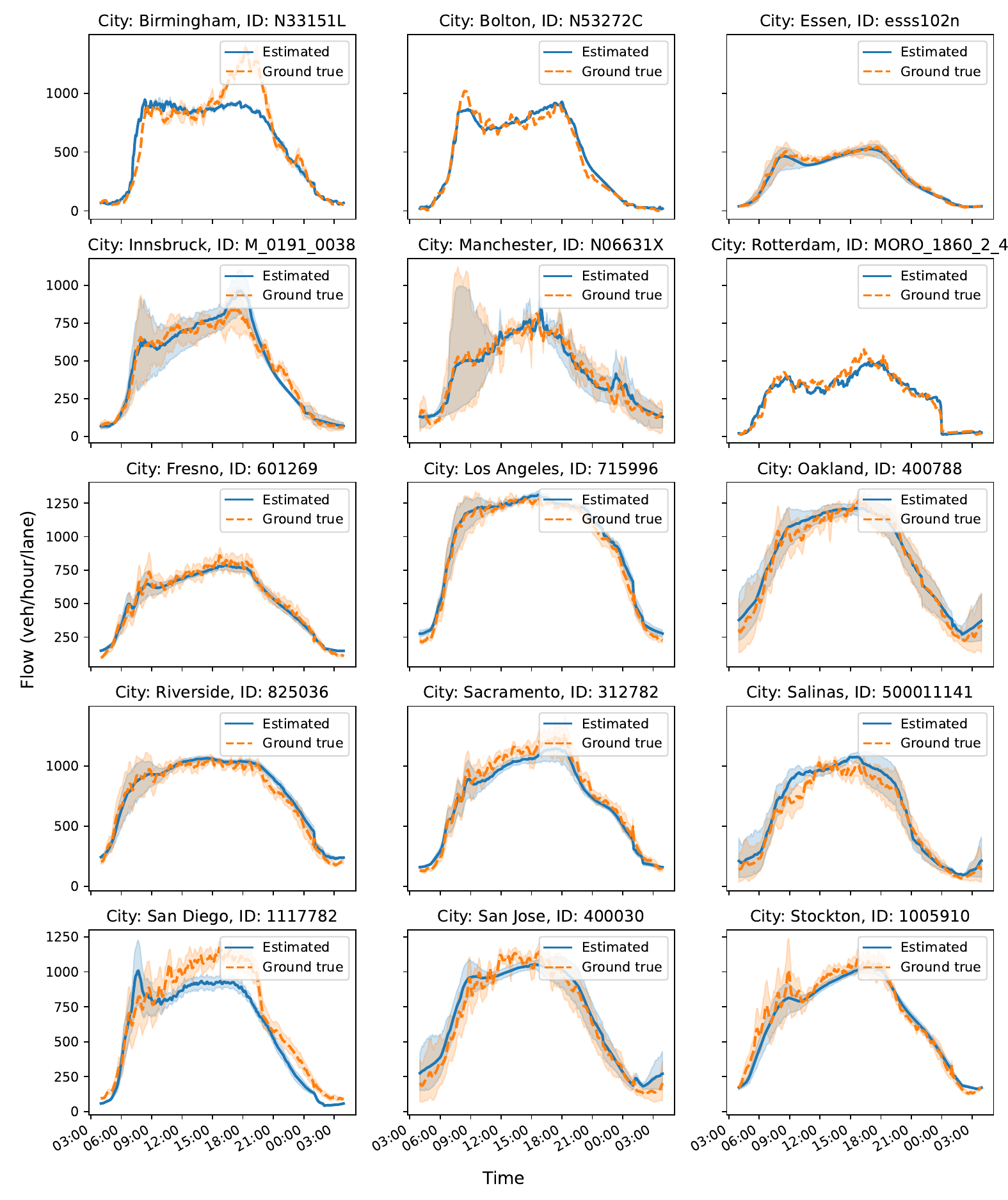}
    \caption{Variation of average daily traffic flow of unseen sensors from the proposed method and the ground truth.}
    \label{fig:ground_truth_line}
\end{figure}

\begin{figure}[ht]
    \centering
    \includegraphics[width=0.90\textwidth]{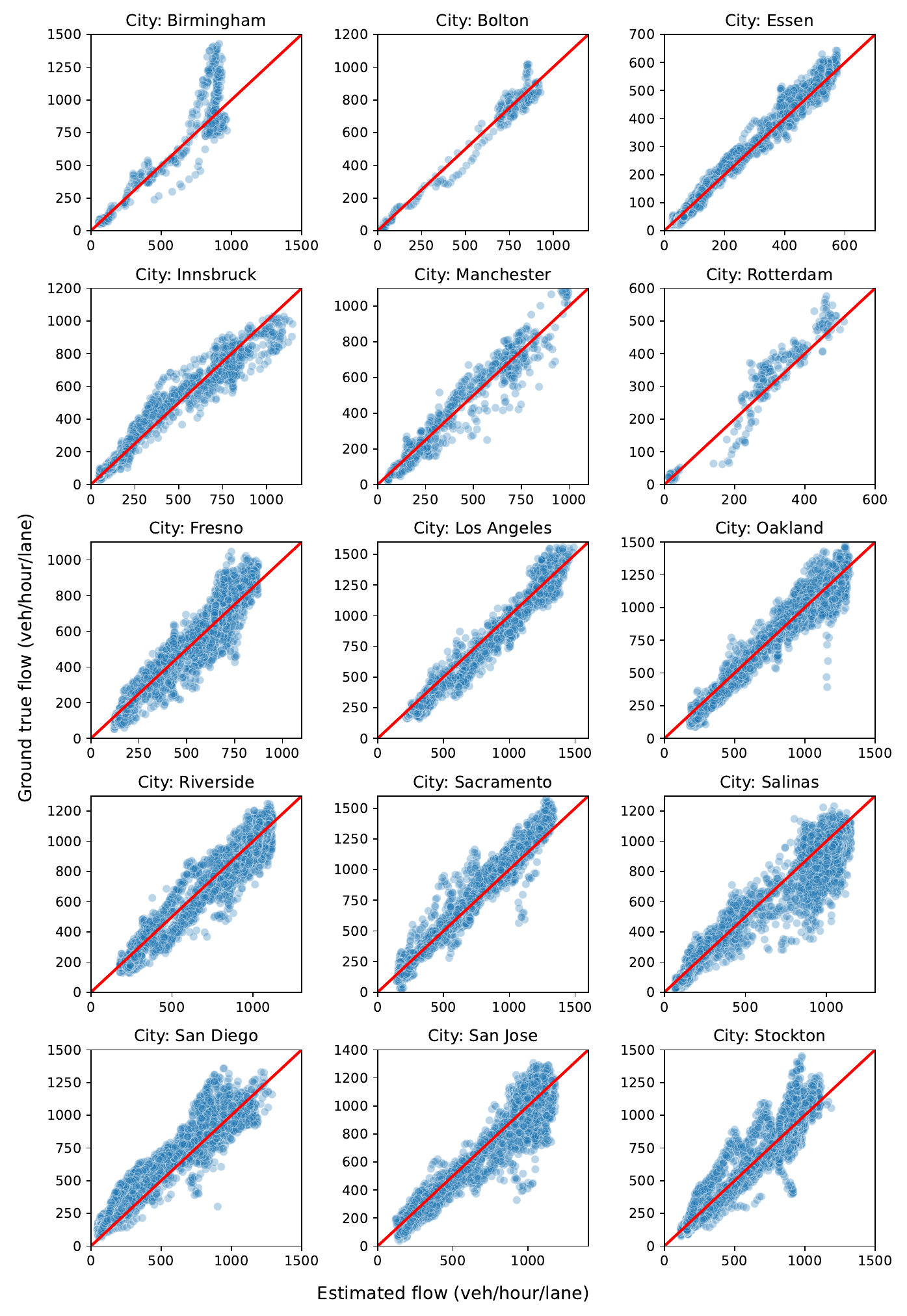}
    \caption{Correlation of the estimated flow with the ground truth. If points are close to the red lines, the estimations are accurate.}
    \label{fig:ground_truth_scatter}
\end{figure}

Furthermore, we compare the estimated flow with the ground truth of all cities in Europe and North America as shown in Figures ~\ref{fig:ground_truth_line} and ~\ref{fig:ground_truth_scatter}. Figure~\ref{fig:ground_truth_line} shows the variation of average daily traffic flow of unseen sensors from the proposed method and the ground truth. Since there are multiple unseen sensors in different cities, and it is difficult to present them all, we randomly select one unseen sensor in each city for comparison. It can be seen that in most cities, the estimation flow variation curve is close to the ground truth, meaning the estimation is accurate, except for Birmingham and San Diego. In Birmingham, the proposed method fails to estimate a flow surge in the evening peak. In San Diego, the flows are underestimated most of the time compared with the ground truth. The discrepancies may happen when the training data is not enough resulting in overfitting in the estimation method, or the relationship between the traffic flow with traffic-related data on unseen roads is significantly different from that in the training set. Overall, the proposed method can achieve an accurate estimation of traffic flow in most cities.

A full comparison of estimation results with the ground truth are shown in Figure~\ref{fig:ground_truth_scatter}. While the the X-axis represents the estimated traffic flow (unit: veh/hour/lane), the Y-axis indicates the value of the ground truth (units: veh/hour/lane). Each point in the figure denotes a testing data point from unseen sensors in the testing set. The red lines show values where the estimated flow is equal to the ground truth, which means the estimated traffic flow is accurate if a point is close to the red line. In this figure, we can see that data points are relatively close to the red lines in most cities, except Birmingham, Stockton, and Salinas. In Birmingham, most data points with estimated values between 600 to 1000 are higher than the red line, indicating an under-estimation of the traffic flow, which matches the results in Figure~\ref{fig:ground_truth_line}. In Stockton, numerous data points are also located above the red line. In Salinas, the data points around 1000 (X-axis) are more scattered than the data points around 500 (X-axis), indicating an inaccurate estimation with higher traffic flow potentially at some specific sensors.

\begin{table}[h]
    \centering
    \caption{NTFE results of the proposed and baseline methods within cities in Europe. For each method, there are three kinds of errors, RMSE, MAE and SMAPE (from top to bottom), respectively. The units for RMSE and MAE are veh/hour/lane.}
    \begin{tabular}{p{0.10\textwidth}|p{0.12\textwidth}p{0.12\textwidth}p{0.12\textwidth}p{0.12\textwidth}p{0.12\textwidth}p{0.12\textwidth}}
    \hline
    Methods & Birmingham & Bolton & Essen & Innsbruck & Manchester & Rotterdam \\
    \hline
    \multirow{3}{*}{SA} 
    & 298.50	& 303.06	& 86.24	& 165.08	& 284.66	& 106.09 \\
    & 263.77	& 246.02	& 61.53	& 136.80	& 218.35	& 77.91 \\
    & 42.98\%	& 79.99\%	& 19.15\%	& 49.44\%	& 50.36\%	& 31.84\% \\
    \hline
    \multirow{3}{*}{LSTM} 
    & 199.39	& 162.05	& 119.03	& 253.93	& 186.21	& 100.10 \\
    & 165.19 	& 125.40	& 82.96 	& 196.37	& 132.87	& 73.21 \\
    & 30.63\%	& 32.22\%	& 29.23\%	& 34.09\%	& 28.96\%	& 31.33\% \\
    \hline
    \multirow{3}{*}{GCMC}
    & 904.81	& 419.88	& 290.45	& 661.60	& 392.79	& 125.54 \\
    & 608.25	& 338.80	& 225.30	& 473.40	& 295.56	&83.17 \\
    & 127.00\%	& 80.98\%	& 89.35\%	& 147.20\%	& 56.54\%	& 42.62\% \\
    \hline
    \multirow{3}{*}{GATv2} 
    & 178.16	& 163.45	& 100.23	& 231.84	& 218.51	& 79.99 \\
    & 149.20	& 127.46	& 71.50 	& 154.15	& 157.02	& 57.72 \\
    & 29.35\%	& 3303\%	& 26.10\%	& 34.17\%	& 34.31\%	& 28.79\% \\
    \hline
    \multirow{3}{*}{\textbf{Ours}} 
    & \textbf{137.62}	& \textbf{100.51}	& \textbf{46.77}	& \textbf{122.38}	& \textbf{161.41}	& \textbf{78.61} \\
    & \textbf{106.21}	& \textbf{74.37}	& \textbf{35.13}	& \textbf{90.21}	& \textbf{113.28}	&  \textbf{55.61} \\
    & \textbf{21.55\%}	& \textbf{23.96\%}	& \textbf{16.79\%}	& \textbf{20.40\%}	& \textbf{23.95\%}	& \textbf{26.29\%} \\
    \hline
    \end{tabular}
    \label{tab:benchmark_eu}
\end{table}

\begin{table}[h]
    \centering
    \caption{NTFE results of the proposed and benchmark methods within cities in North America. For each method, there are three kinds of errors, RMSE, MAE and SMAPE (from top to bottom), respectively. The units for RMSE and MAE are veh/hour/lane.}
    \begin{tabular}{p{0.07\textwidth}|p{0.05\textwidth}p{0.10\textwidth}p{0.07\textwidth}p{0.07\textwidth}p{0.09\textwidth}p{0.055\textwidth}p{0.08\textwidth}p{0.07\textwidth}p{0.07\textwidth}}
    \hline
    Methods & Fresno & Los Angeles & Oakland & Riverside & Sacramento & Salinas & San Diego & San Jose & Stockton \\
    \hline
    \multirow{3}{*}{SA} 
    & 254.91	& 416.64	& 254.88	& 321.62	& 293.58	& 286.16	& 195.91	& 230.83	& 297.88 \\
    & 167.21	& 312.71	& 185.35	& 248.36	& 217.27	& 200.97	& 146.15	& 169.26	& 210.22 \\
    & 28.40\%	& 48.35\%	& 26.18\%	& 38.06\%	& 28.15\%	& 42.63\%	& 24.97\%	& 29.12\%	& 33.35\% \\
    \hline
    \multirow{3}{*}{LSTM} 
    & 256.77	& 258.38	& 237.41	& 229.10	& 242.19	& 140.48	& 208.25	& 192.95	& 206.43 \\
    & 152.51	& 176.98	& 176.99	& 165.30	& 162.43	& 99.86 	& 148.19	& 139.55	& 143.70 \\
    & 31.54\%	& 25.05\%	& 26.18\%	& 33.20\%	& 23.61\%	& 29.87\%	& 27.63\%	& 25.93\%	& 25.42\% \\
    \hline
    \multirow{3}{*}{GCMC} 
    & 317.07	& 381.46	& 849.64	& 315.63	& 370.42	& 345.11	& 373.50	& 467.98	& 338.84 \\
    & 199.04	& 223.40	& 705.83	& 253.02	& 281.29	& 268.87	& 314.01	& 391.45	& 221.03 \\
    & 31.57\%	& 24.54\%	& 174.50\%	& 37.91\%	& 32.81\%	& 43.33\%	& 33.07\%	& 70.46\%	& 43.59\% \\
    \hline
    \multirow{3}{*}{GATv2}
    & 249.45	& 291.55	& 226.07	& 238.89	& 230.22	& 141.80	& 218.75	& 183.95	& 191.90 \\
    & 146.95	& 222.38	& 166.24	& 178.34	& 173.21	& 100.26	& 162.09	& 132.83	& 134.71 \\
    & 30.60\%	& 29.90\%	& 24.34\%	& 35.29\%	& 24.41\%	& 30.87\%	& 28.50\%	& 24.89\%	& 24.19\% \\
    \hline
    \multirow{3}{*}{Ours} 
    & \textbf{160.21}	& \textbf{219.35}	& \textbf{208.72}	& \textbf{209.62}	& \textbf{201.81}	& \textbf{140.42}	& \textbf{173.36}	& \textbf{180.76}	& \textbf{169.08} \\
    & \textbf{114.21}	& \textbf{160.49}	& \textbf{146.93}	& \textbf{147.16}	& \textbf{143.07}	& \textbf{97.63}	& \textbf{124.80}	& \textbf{129.67}	& \textbf{126.63} \\
    & \textbf{21.29\%}	& \textbf{23.65\%}	& \textbf{21.32\%}	& \textbf{26.78\%}	& \textbf{21.68\%}	& \textbf{26.93\%}	& \textbf{23.90\%}	& \textbf{23.93\%}	& \textbf{19.78\%} \\
    \hline
    \end{tabular}
    \label{tab:benchmark_na}
\end{table}

\clearpage
\subsection{Ablation study} \label{sec:abl_study}
In this section, we conduct two ablation studies to validate: 1) whether the GOMS is crucial to the NTFE; 2) whether the proposed TCAB and DCB can extract useful information from the GOMS maps.

\subsubsection{Ablation study on data sources}
In this study, an important question is whether the utilization of GOMS maps is necessary for the NTFE. To answer the question, we conduct an ablation study on data sources. We inspect how the method performance changes if we exempt part of the input data. The experiments are separated into 8 groups. We use no map data, use only OSM, we use only sensor distribution map, we use only population density map, we use both OSM and sensor distribution map, we use both OSM and population density map, we use both sensor distribution map and population density map, and we use all data. The results of each group across the 15 cities are shown in Table~\ref{tab:as_data_eu} and ~\ref{tab:as_data_na}.
\begin{table}[ht]
    \centering
    \caption{The results of ablation studies on data sources within cities in Europe. For each method, there are three kinds of errors which are RMSE, MAE and SMAPE (from top to bottom), respectively. The unit for RMSE and MAE is veh/hour/lane.}
    \begin{tabular}{p{0.10\textwidth}|p{0.12\textwidth}p{0.12\textwidth}p{0.12\textwidth}p{0.12\textwidth}p{0.12\textwidth}p{0.12\textwidth}}
    \hline
    Methods & Birmingham & Bolton & Essen & Innsbruck & Manchester & Rotterdam \\
    \hline
    \multirow{3}{*}{None} 
    & 185.16	& 135.84	& 101.22	& 247.27	& 185.73	& 107.75 \\
    & 138.70	& 100.92	& 71.37 	& 181.76	& 134.69	& 77.27  \\
    & 26.97\%	& 30.49\%	& 27.23\%	& 34.85\%	& 28.60\%	& 32.56\% \\
    \hline
    \multirow{3}{*}{OSM only} 
    & 169.71	& 123.93	& 70.57 	& 136.68	& 191.45	& 109.98 \\
    & 133.67	& 95.95 	& 49.41 	& 104.04	& 133.18	& 80.28  \\
    & 25.57\%	& 28.84\%	& 21.67\%	& 22.82\%	& 27.49\%	& 35.51\% \\
    
    \hline
    \multirow{3}{*}{Loc only} 
    & 179.73	& 122.51	& 60.80 	& 132.56	& 191.14	& 125.27 \\
    & 148.14	& 80.37 	& 46.32 	& 96.40 	& 134.76	& 93.11  \\
    & 28.61\%	& 29.41\%	& 21.40\%	& 21.50\%	& 28.74\%	& 34.51\% \\
    \hline
    \multirow{3}{*}{Pop only} 
    & 208.75	& 127.14	& 75.87 	& 130.97	& 183.24	& 133.15 \\
    & 159.35	& 90.40 	& 53.51 	& 93.91 	& 131.00	& 99.24  \\
    & 26.28\%	& 28.34\%	& 23.13\%	& 21.19\%	& 27.93\%	& 40.13\% \\
    \hline
    \multirow{3}{*}{OSM+Loc} 
    & 175.76	& 104.11	& 54.06 	& 133.00	& 170.94 	& 116.53 \\
    & 133.73	& 82.33 	& 39.39 	& 89.64 	& 116.55 	& 85.67 \\
    & 22.33\%	& 24.29\%	& 18.52\%	& 21.22\%	& 24.39\%	& 33.56\% \\
    \hline
    \multirow{3}{*}{OSM+Pop} 
    & 188.62	& 101.24	& 53.51 	& 160.44 	& 173.54	& 106.40 \\ 
    & 147.75	& 75.42 	& 38.36 	& 119.54 	& 118.82	& 79.57  \\
    & 26.58\%	& 24.64\%	& 17.15\%	& 22.63\%	& 24.90\%	& 34.10\%  \\
    \hline
    \multirow{3}{*}{Loc+Pop} 
    & 155.03	& 107.07	& 51.55 	& \textbf{95.42} 	& 184.45	& 116.63 \\
    & 119.62	& 76.04 	& 37.08 	& \textbf{68.09} 	& 127.80	& 84.29  \\
    & 22.53\%	& 24.68\%	& 17.62\%	& \textbf{17.08\%}	& 26.78\%	& 34.54\%  \\
    \hline
    \multirow{3}{*}{All} 
    & \textbf{137.62}	& \textbf{100.51}	& \textbf{46.77} 	& 122.38	& \textbf{161.41} 	& \textbf{78.61}  \\
    & \textbf{106.21}	& \textbf{74.37} 	& \textbf{35.13} 	& 90.21 	& \textbf{113.28} 	& \textbf{55.61}  \\
    & \textbf{21.55\%}	& \textbf{23.96\%}	& \textbf{16.79\%}	& 20.40\%	& \textbf{23.95\%}	& \textbf{26.29\%} \\
    \hline
    \end{tabular}
    \label{tab:as_data_eu}
\end{table}
\begin{table}[ht]
    \centering
    \caption{Results of ablation studies on data sources within cities in North America. For each method, there are three kinds of errors, RMSE, MAE and SMAPE (from top to bottom), respectively. The units for RMSE and MAE are veh/hour/lane.}
    \begin{tabular}{p{0.08\textwidth}|p{0.05\textwidth}p{0.10\textwidth}p{0.07\textwidth}p{0.07\textwidth}p{0.09\textwidth}p{0.055\textwidth}p{0.08\textwidth}p{0.07\textwidth}p{0.07\textwidth}}
    \hline
    Methods & Fresno & Los Angeles & Oakland & Riverside & Sacramento & Salinas & San Diego & San Jose & Stockton \\
    \hline
    \multirow{3}{*}{None} 
    & 256.49 & 297.21 & 224.10 & 225.74 & 219.59 & 144.16 & 224.46 & 188.99 & 195.50  \\
    & 148.00 & 226.43 & 168.96 & 158.79 & 159.60 & 100.04 & 163.27 & 136.75 & 134.94  \\
    & 29.98\% & 30.56\% & 24.49\% & 31.99\% & 23.23\% & 30.35\% & 27.61\% & 25.01\% & 24.23\%  \\
    
    \hline
    \multirow{3}{*}{OSM only} 
    & 252.29	& 244.62	& 216.45	& 222.54	& 215.79	& \textbf{139.24}	& 193.94	& 190.26	& 199.41 \\
    & 147.73	& 188.37	& 147.97	& 159.22	& 156.40	& 98.66 	& 141.58	& 133.59	& 143.14 \\
    & 29.44\%	& 24.85\%	& 21.37\%	& 32.55\%	& 23.04\%	& 28.29\%	& 26.44\%	& 24.43\%	& 24.84\% \\
    
    \hline
    \multirow{3}{*}{Loc only} 
    & 255.56	& 246.99	& 240.81	& 228.05	& 202.51	& 146.22	& 190.12	& 189.05	& 220.94 \\
    & 159.35	& 180.47	& 175.59	& 163.53	& 146.46	& 100.95	& 139.60	& 137.87	& 159.35 \\
    & 27.98\%	& 25.91\%	& 24.60\%	& 29.29\%	& 22.74\%	& 28.31\%	& 26.17\%	& 25.24\%	& 27.36\% \\
    
    \hline
    \multirow{3}{*}{Pop only} 
    & 258.98	& 348.54	& 234.50	& 233.53	& 224.04	& 145.85	& 196.71	& 194.44	& 214.47 \\
    & 153.49	& 228.57	& 175.02	& 167.72	& 160.52	& 100.79	& 146.10	& 140.71	& 155.37 \\
    & 27.41\%	& 27.09\%	& 24.86\%	& 29.76\%	& 23.65\%	& 27.77\%	& 28.02\%	& 25.47\%	& 26.71\% \\
    
    \hline
    \multirow{3}{*}{OSM+Loc} 
    & 216.78	& 224.32	& 212.33	& 224.99	& 202.64	& 141.21	& 176.00	& 181.14	& 188.71 \\
    & 133.20	& 163.92	& 155.37	& 164.62	& 145.32	& 98.20 	& 127.09	& 130.38	& 132.78 \\
    & 23.71\%	& 24.31\%	& 21.50\%	& 28.99\%	& \textbf{20.41\%}	& 27.15\%	& 24.71\%	& 24.56\%	& 23.52\% \\
    
    \hline
    \multirow{3}{*}{OSM+Pop} 
    & 223.56	& 226.32	& 215.53	& \textbf{206.12}	& 201.81	& 146.58	& 182.60	& 185.20	& 193.52 \\
    & 142.90	& 165.92	& 155.88	& 152.25	& \textbf{142.25}	& 102.64	& 132.75	& 134.02	& 140.17 \\
    & 25.18\%	& 24.47\%	& 22.02\%	& 28.47\%	& 21.33\%	& \textbf{26.25\%}	& 24.81\%	& 24.86\%	& 24.49\% \\
    
    \hline
    \multirow{3}{*}{Loc+Pop} 
    & 244.07	& 228.02	& 210.15	& 218.64	& 204.49	& 139.70	& 175.61	& 180.78	& 215.30 \\
    & 145.32	& 165.24	& 151.92	& 155.07	& 145.51	& 98.81 	& 128.76	& 131.93	& 154.35 \\
    & 25.61\%	& 24.63\%	& 21.66\%	& 28.90\%	& 22.52\%	& 27.10\%	& 24.14\%	& 24.31\%	& 26.71\% \\
    
    \hline
    \multirow{3}{*}{All} 
    & \textbf{160.90}	& \textbf{219.35}	& \textbf{208.72}	& 209.62	& \textbf{199.21}	& 140.42	& \textbf{173.36}	& \textbf{180.76}	& \textbf{169.08} \\
    & \textbf{114.21}	& \textbf{160.49}	& \textbf{146.93}	& \textbf{147.16}	& 143.07	& \textbf{97.63} 	& \textbf{124.80}	& \textbf{129.67}	& \textbf{126.63} \\
    & \textbf{21.29\%}	& \textbf{23.65\%}	& \textbf{21.32\%}	& \textbf{26.78\%}	& 21.68\%	& 26.93\%	& \textbf{23.90\%}	& \textbf{23.93\%}	& \textbf{19.78\%} \\
    \hline
    \end{tabular}
    \label{tab:as_data_na}
\end{table}
It can be seen that the estimation tends to become increasingly accurate as more GOMS maps are incorporated into the method. When only the public traffic data and network data are utilized, the estimation results are not decent. However, the accuracy gradually increases as we incorporate more data. In most cities, the errors reach the minimum when all geographical and demographical data are incorporated in the estimation method. 

\begin{figure}
    \centering
    \includegraphics[width=0.95\textwidth]{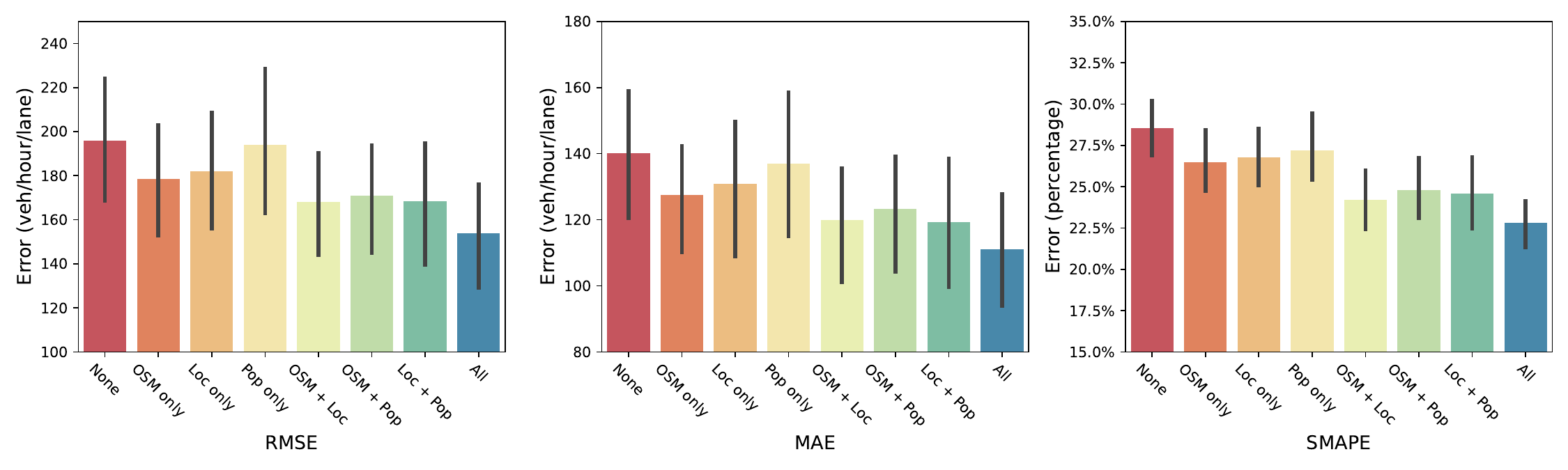}
    \caption{Results of ablation studies on data sources within all cities. The black lines represent the standard deviation.}
    \label{fig:ablation_image}
\end{figure}

A summary of estimation errors in all cities with different data sources is shown in Figure~\ref{fig:ablation_image}.
The three subplots represent the average errors (RMSE, MAE and SMAPE, respectively), the black line in each subplot denotes the standard deviation. It can be seen that, if we only incorporate one image in the estimation method, the method with only OSM performs best. The probable reason is the OSM contains affluent spatial geographical information that can contribute to the flow estimation. The method with only population map performs worst since the included geographical and demographical information may be limited. Incorporating two images further reduces the estimation error, but the performance varies according to different error metrics.

Importantly, the proposed estimation method is flexible to add or replace some data sources. With the development of sensor technologies, it is foreseeable that more GOMS maps will emerge with the advent of smart cities and large foundation methods. Consequently, more GOMS maps can be applied in the future to increase the estimation accuracy.

\subsubsection{Ablation study on network components}
The validity of the different network components is another important question to discuss. In this study, we propose two novel neural network blocks, TCAB and DCB, to extract the information from the utilized multi-source data. Besides, we use both GSAM and GTAM to encode the spatial and temporal characteristics in graph data. The impact of each block on the estimation accuracy will be validated in this section.

\begin{figure}[h]
    \centering
    \includegraphics[width=0.95\textwidth]{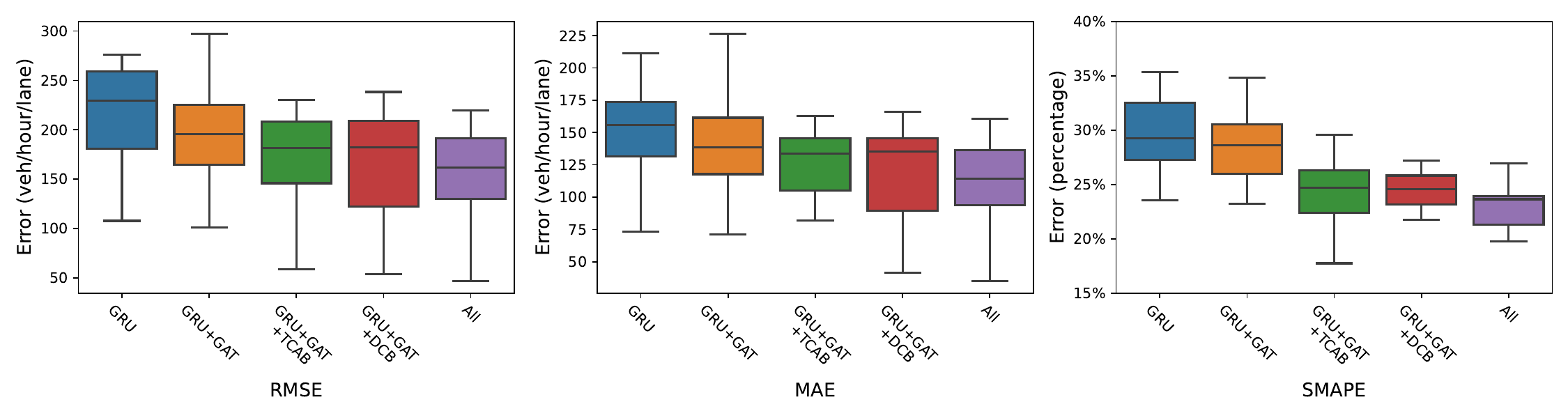}
    \caption{The boxplots of estimation errors of ablated components within all cities.}
    \label{fig:ablation_com}
\end{figure}

We divide the ablation study into 5 groups. We only use GRU (model temporal characteristics), we use both GRU and GAT (model both spatial and temporal characteristics), we use GRU, GAT and TCAB, we use GRU, GAT and DCB, and we use all blocks. The estimation results for each group across all cities in Europe and North America are presented in Figure~\ref{fig:ablation_com}. 
One can see that the estimation accuracy is higher as we utilize more components in the neural network. The maximal, minimal and median errors are reduced when we incorporate the TCAB or DCB. The estimation error is further reduced when we incorporate all blocks in the neural network.

\subsection{Sensitivity analysis} \label{sec:sen_analysis}
In this section, we conduct three sensitivity analyses from different aspects. We mainly focus on two representative cities: Manchester in Europe and Oakland in North America, respectively. The analyses in this section are all carried out for these two cities.

\subsubsection{Sensitivity analysis of training set coverage}
Since the realm and topology of cities are different, the number of nodes in all cities is different in this study. An interesting question is whether the node number in the training set will affect the final estimation result. To answer this question, we conduct a sensitivity analysis on training set coverage. Specifically, we randomly select a proportion of nodes in the training set with different ratios and make them new training sets, while keeping the same testing set. The ratios are set to 25\%, 50\%, 75\% and 100\%. For example, if the ratio equals 25\%, the number of nodes in the training set is 25\% of the original training set, while the testing set is unchanged. We anticipate that reducing the size of the training set may decrease the performance of the method. Each of these new training sets is then used to individually train neural networks, which are evaluated on the same unchanged testing set.

\begin{figure}[h]
    \centering
    \includegraphics[width=0.95\textwidth]{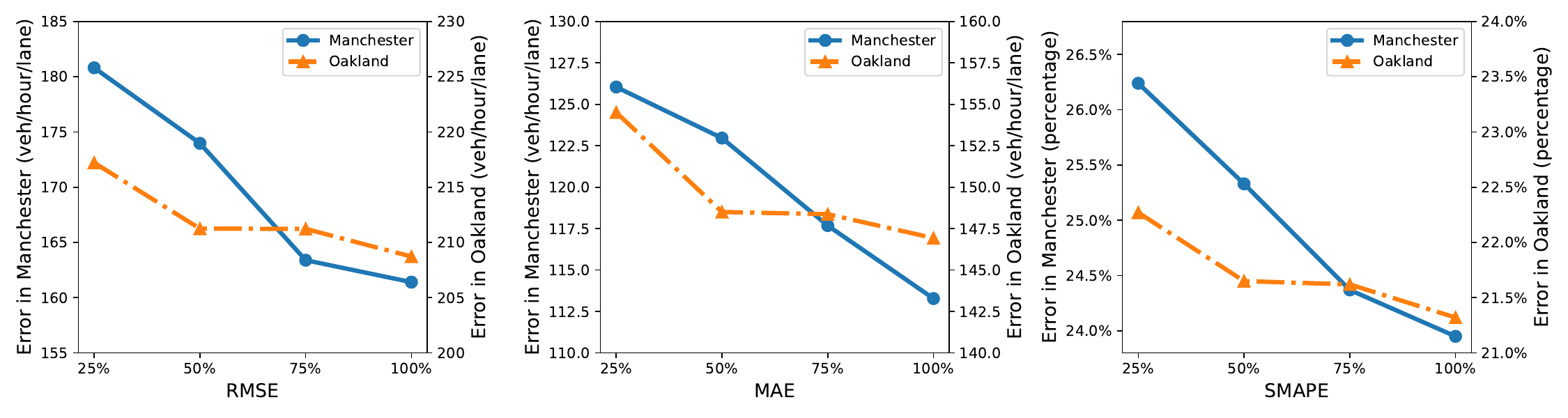}
    \caption{The estimation errors of our method on different training sets in Manchester and Oakland. The X-axis denotes the proportion of selected nodes in the original training set. The left and right Y-axes mean errors in Manchester and Oakland separately.}
    \label{fig:sensitivity_coverage}
\end{figure}

The results are shown in Figure~\ref{fig:sensitivity_coverage}. Since the estimation errors range differently in Manchester and Oakland, we fix the same scale for the same error metric. In both Manchester and Oakland, the estimation error decreases with more nodes in the training set. However, in Oakland, the error decreases minorly from 25\% training set to 100\% training set. The SMAPE only decreases less than 1\% while the RMSE decreases less than 10 veh/hour/lane. The performance of neural networks on 50\% and 75\% training sets are nearly the same. In Manchester, the decrease is more significant. The SMAPE decreases about 2.5\% from 25\% training set to 100\% training set, and the MAE decreases about 20 veh/hour/lane. 

The reason for different estimation errors in Manchester and Oakland could be the absolute number of road segments in the training set. There are 204 road segments in Oakland and 122 road segments in Manchester in the training set. If we keep 25\% and 50\% road segments in Oakland, there will be 51 and 102 roads in the new training set. Meanwhile, if we keep 50\% and 100\% road segments in Manchester, there will be 61 and 122 in the new training set. The RMSE and SMAPE of groups 25\% and 50\% training set in Oakland approaches the RSME and SMAPE of groups 50\% and 100\% training set in Manchester. Therefore, the absolute number of road segments in the training set might be more important to the performance of the proposed method. In a very instrumented city, even if we keep a small proportion of road segments for training, the estimation error may still be decent. However, in a city with sparse sensors, we need a large proportion of all sensors for training to reach a decent performance.

\subsubsection{Sensitivity analysis of testing set selection}
In the previous sensitivity analysis, we discuss the impact of the training set coverage on the estimation performance. The selection of the testing set is still important to the justice of evaluation. In this sensitivity analysis, we discuss the impact of testing set selection on the performance. We randomly split training nodes and testing nodes in Manchester and Oakland 10 times, where the training and testing node numbers are the same as the numbers in Table~\ref{tab:datasets}. Then, we individually train the neural network on a training set and evaluate it on a testing set. 

\begin{figure}[h]
    \centering
    \includegraphics[width=0.95\textwidth]{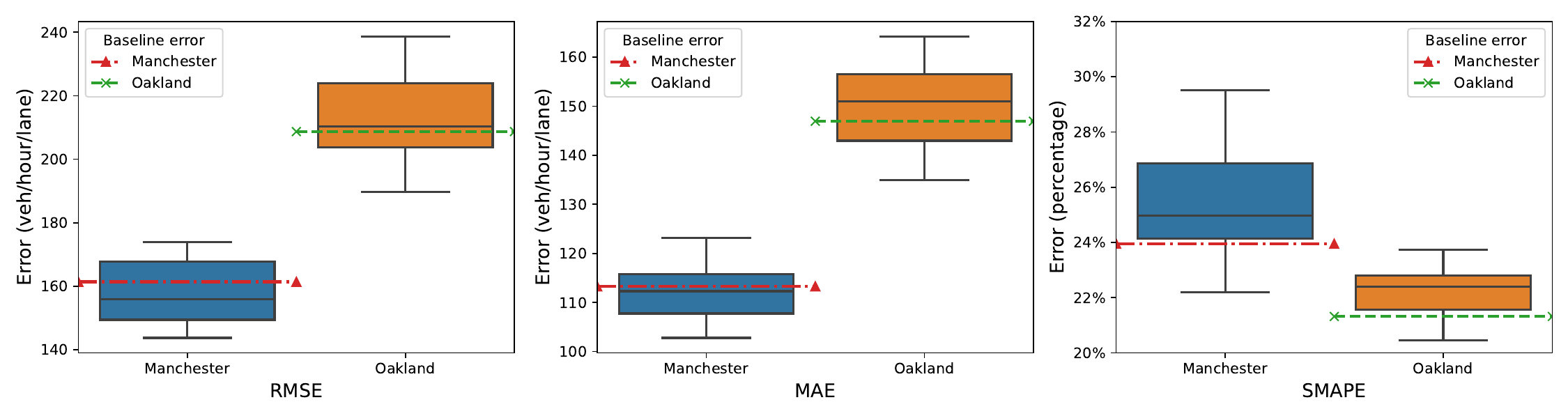}
    \caption{The boxplots of estimation errors with different training and testing sets in Manchester and Oakland. In each subplot, the red and green dashed lines represent the baseline errors in Manchester and Oakland in Table~\ref{tab:benchmark_eu} and~\ref{tab:benchmark_na}.}
    \label{fig:sensitivity_selection}
\end{figure}

The estimation errors are shown in Figure~\ref{fig:sensitivity_selection}. From the perspective of RMSE and MAE, the baseline errors are located between the first quartile to the third quartile in both Manchester and Oakland, while the baseline SMAPEs are smaller than the first quartiles but larger than the minimums in both cities. Moreover, the interquartile range of SMAPE in Manchester is larger than the one in Oakland, indicating that the method performs more robustly in Oakland. The selection of the testing set indeed impacts the estimation accuracy, but such an impact decreases with the growth of sensor number.


\section{Conclusion}\label{sec:conclusion}
In this study, we conduct a large-scale case study for NTFE in Europe and North America by leveraging the GOMS maps in a deep learning framework, which is the first of its kind in the literature.
Specifically, the GOMS maps consist of three maps that can be publicly accessed: OSM, sensor distribution map, and population density map. Compared with tabular data, such map images not only contain fruitful geographical and demographical information that can contribute to the NTFE but also give rise to a unified data format. Moreover, we develop an attention-based deep-learning method to fully exploit the geographical and demographical information from GOMS maps. The novel triple cross-attention and dense connection blocks have been developed to extract information from GOMS maps.
Experimental results demonstrate that the utilization of GOMS maps can indeed fulfill both general and accurate NTFE in multiple cities. The ablation study also demonstrates that utilizing comprehensive and unified GOMS maps provides a significant improvement over traditional supplementary data. The GOMS maps offer a more detailed and global collection of information, which enhances the accuracy and generality of the NTFE results. 

In future research, more extensions regarding the data sources can be made based on the current study. With the development of sensor technologies, it is expected that more and more GOMS data, such as meteorological data, will become available, and utilizing more diverse GOMS data could further enhance the estimation accuracy. Moreover, we currently train a separate neural network for each city to guarantee the estimation accuracy. However, our goal is to develop a universal method that performs well in multiple cities without requiring additional training, potentially based on Large Foundation Models.  Furthermore, we plan to collect data from additional cities in China, Japan, and other countries to evaluate our method on a global scale.

\section*{Acknowledgments}
The work described in this paper is supported by grants from the Research Grants Council of the Hong Kong Special Administrative Region, China (Project No. PolyU/15206322 and PolyU/15227424). M. Menendez acknowledges the support of the NYUAD Center for Interacting Urban Networks (CITIES), funded by Tamkeen under the NYUAD Research Institute Award CG001. The contents of this article reflect the views of the authors, who are responsible for the facts and accuracy of the information presented herein. 

\printcredits

\bibliographystyle{apalike}  

\bibliography{main}



\end{document}